\documentclass[letterpaper,12pt]{article} 
\usepackage{amsmath} 
\usepackage[american]{babel}
\usepackage{tabularx}
\usepackage{booktabs}
\usepackage{graphicx}

\pdfoutput=1 
\title{Derivation of the Variational Bayes Equations}

\author{Alianna J. Maren \\
  Themesis Technical Report TR-2019-01v6 (ajm)\\ 
  {\tt themesis1@themesis.com}\\
  {\tt alianna.maren@northwestern.edu}
}

\begin{document}
%

\maketitle


\abstract{The derivation of key equations for the variational Bayes approach is well-known in certain circles. However, translating the fundamental derivations (e.g., as found in Beal's work) to Friston's notation is somewhat delicate. Further, the notion of using variational Bayes in the context of a system with a Markov blanket requires special attention. This Technical Report presents the derivation in detail. It further illustrates how the variational Bayes method provides a framework for a new computational engine, incorporating the 2-D cluster variation method (CVM), which provides a necessary free energy equation that can be minimized across both the external and representational systems' states, respectively. }


\begin{flushright}
``Do you understand this?'' she demanded. ... \\

``It seems simple enough,'' he said after a moment.\\

``I knew it,'' she muttered, crossing her arms. ``I knew it was written in male.''\\

\textit{Heir to the Shadows: Book 2 of the Black Jewels Trilogy}\\

Anne Bishop (1999), p. 214 (Trade edition). 
 \end{flushright}

\newcommand\T{\rule{0pt}{2.6ex}}       
\newcommand\B{\rule[-1.2ex]{0pt}{0pt}} 


%
\section{Introduction}
\label{sec:Introduction}
%

A recent evolution by Friston et al. (2024) \cite{Friston-et-al_2024_From-pixels-to-planning} illustrates how active inference can be used for multi-scale applications, which in turn prompts greater attention to the derivation of core active inference equations. 

Beyond this, the recent exposition by Friston et al. (2023) \cite{Friston-et-al_2023_Free-energy-principle} on the ``free energy principle,'' coupled with a conceptual advance by Hafner et al. (2020, rev. 2022) \cite{Hafner-et-al_2022_Action-perception-divergence} on``Action Perception Divergence" (APD) posits the need to trace how the active inference equations have been derived, as well as adapted for expression in different circiumstances. 

Friston (2010, 2013) \cite{Friston_2010_Free-energy-principle-unified-brain-theory, Friston_2013_Life-as-we-know-it} has proposed that free energy minimization serves as a unifying theory for describing neural dynamics, with further elaboration in Friston et al. (2015) \cite{Friston-et-al_2015_Knowing-ones-place-free-energy-pattern-recognition}. He further suggests that  statistical thermodynamics can model neuronal systems, drawing on the dynamic properties of activated neuronal ensembles \cite{Friston-et-al_2015_Knowing-ones-place-free-energy-pattern-recognition}. This elegant and fascinating notion depends on the use of the variational Bayes approach together with the idea of a Markov blanket, which separates an internal computational (``representational'') set of units from external ones.

While this approach is certainly attractive, it is potentially difficult for many readers to follow the translation from one of the earlier presentations of variational Bayes (by Beal (2003) \cite{Beal_2003_Variational-algorithm-approx-Bayes-inference}) to the equations used by Friston (op. cit.). 

The treatment offered in this work is a careful deconstruction of the ideas and formalism as they were originally articulated by Friston (op. cit.) based on detailed derivations used by Beal \cite{Beal_2003_Variational-algorithm-approx-Bayes-inference}).  As such, it serves as a \textit{Rosetta stone}, not just for the ideas, but also for the meaning of variables and operators as used in different works. This is not as easy as it may seem; for example, the term $H$ could be read as enthalpy in thermodynamic treatments, while it stands in for entropy in purely information theoretic treatments. 

Thus, this Technical Report serves as a mini-tutorial, carefully delineating how the free energy equations presented in Friston (op. cit.) correspond to the detailed derivations presented in Beal (2003), which were originally presented in Feynman \cite{Feynman_1972_Statistical-Mechanics} and in Hinton and van Camp \cite{Hinton-van-Camp_1993_Keeping-neural-networks-simple}. It identifies how - although the equations may seem formally identical - there are certain key differences in the presentations offered by Friston and Beal.

In what follows, we will address the meaning of each variable explicitly and note any instances of overloading (i.e., the use of the same variable to mean two things). To help in this regard, Section~\ref{sec:var-free-energy} includes a glossary of the thermodynamic variables in this paper, along with a brief description. Section~\ref{sec:important-distinction-and-clarification} includes a glossary of the additional information-theoretic notions. 

Also, since Blei et al. (2016, 2017) \cite{Blei-et-al_2016_Variational-Bayes, Blei-et-al_2017_Variational-Bayes} have offered a valuable and useful tutorial on variational inference - one that is usefully read hand-in-hand with Beal - we also address the nomenclature used by Blei et al. Section~\ref{sec:var-FE-log-likelihood-and-KL-divergence} offers a table that compares (in a ``Rosetta stone" manner) the nomenclature used by Beal, Friston, and Blei et al.  

One of the most important elements in the derivation of variational Bayes is that the fundamental free energy equation (Eqn.~\ref{eqn:var-free-energy-eqn_part2-first-time}) can be expressed two different ways. We work through the derivation of the first version in Section~\ref{sec:var-FE-log-likelihood-and-KL-divergence}, and the second in Section~\ref{sec:var-FE-log-likelihood-expectation-and-entropy}. Section~\ref{sec:discussion}, ``\textit{Discussion},"  offers a contrast-and-compare of these two different free energy expressions. 

Between 2016 and 2017, Friston and colleagues shifted the notation that they used, moving from explicit depiction of an external surround ($\Psi$) to a notation where the interaction between the system that was being modeled and the surround was evidenced by \textit{action agents} and \textit{sensing agents} \cite{Friston-et-al_2016_Active-inference-and-learning, Friston-et-al_2017_Active-inference-process-theory}. (These agents were present in the earlier Friston works, but now the role of  $\Psi$ became more implicit.) 

The expressions used in Friston et al. (2017) became the basis for much future work, including the \textit{free energy principle} exposition \cite{Friston-et-al_2023_Free-energy-principle} as well as Action Perception Divergence (APD) \cite{Hafner-et-al_2022_Action-perception-divergence}. Thus, in this (2024) revised version of this work, we introduce a new Section~\ref{sec:evolution-active-inference}, which overviews how active inference is presented in Friston et al. (2016, 2017) \cite{Friston-et-al_2016_Active-inference-and-learning, Friston-et-al_2017_Active-inference-process-theory}. 

This work also introduces a new computational engine (Maren, 2016) \cite{Maren_2016_CVM-primer-neurosci} in the active inference context. Such an engine would make use of not only Friston's notion of a set of computational (representational) units separated from an external system by a Markov blanket, but also follow the variational Bayes (free energy minimization) approach described by both Friston (op. cit.) and Beal (2003).  

\begin{figure}
\label{fig:computational-engine-1}
  \centering
  \fbox{
  \rule[0cm]{0cm}{0cm}\rule[0cm]{0cm}{0cm}	
  \includegraphics [trim=0.0cm 0cm 0.0cm 0cm, clip=true,   width=0.9\linewidth]{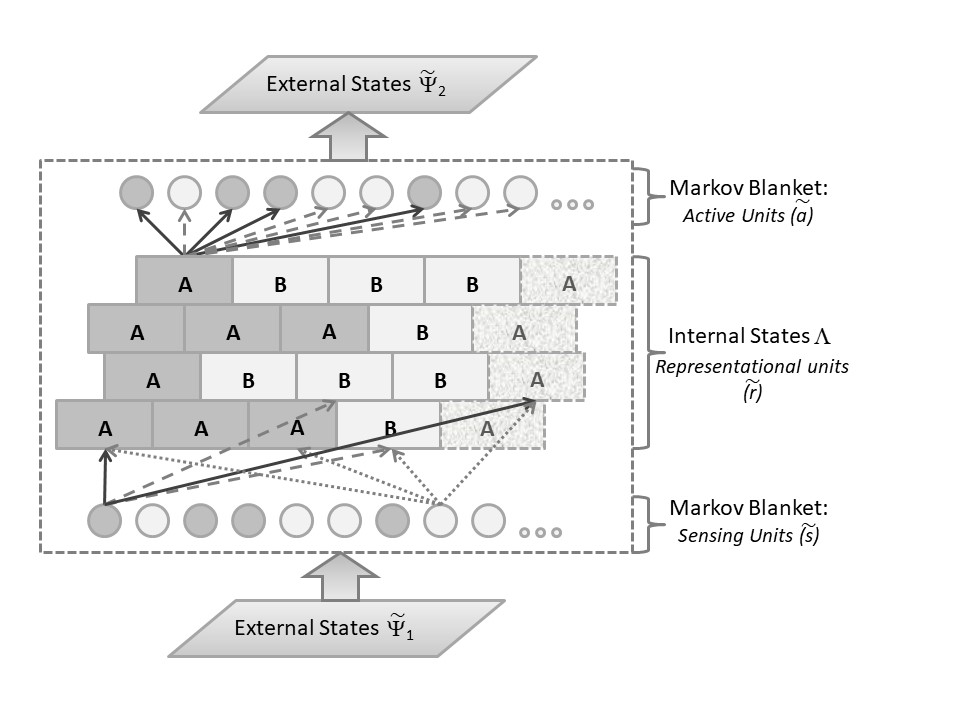}} 
  \vspace{3mm} 
  \caption{Illustration of a CORTECON(R) (COntent-Retentive, TEMporally-CONnected neural network) computational engine (Maren, 2016) \cite{Maren_2016_CVM-primer-neurosci}, which includes an internal latent node grid. Within this grid, the total number of active nodes is govered by an activation enthalpy ($\varepsilon_0$) and the degree of clustering is governed by an interaction enthalpy parameter ($\varepsilon_1$). The  cluster variation method (CVM) is used to bring the active and non-active nodes into free energy equilibrium. A  Markov blanket of sensing and active units corresponds to input and output layers (see Friston \cite{Friston-et-al_2015_Knowing-ones-place-free-energy-pattern-recognition}). The latent node grid, or``computational layer,'' can be composed as either a 1-D or 2-D CVM, for which the free energy minimum can be found either analytically (for the case where $\varepsilon_0 = 0$) or computationally (for the case where $\varepsilon_0 \neq 0$). The CVM layer comprises  the internal or representational units ($\tilde{r}$), and cannot communicate with the external field (shown in two parts for visualization purposes only). However, units within the representational layer can receive inputs from the sensory units ($\tilde{s}$) and send signals to the active ($\tilde{a}$) units. The sensory units can receive inputs from external stimulus, and send signals to the representational units. The active units can receive inputs from the representational units, and send signals to the external system.}
\label{fig:CVM-2-D_comput-engine_crppd_2017-05-13}
\end{figure}

In brief, Friston proposes a computational system in which a Markov blanket separates the computational (representational) elements of the engine from external events, as shown in Figure~\ref{fig:computational-engine-1}. The communications between the external system elements (denoted $\tilde{\psi}$) with those of the representational system (denoted $\lambda$ or $\tilde{r}$) are mediated by two distinct layers or components of the Markov blanket; the sensing ($\tilde{s}$) elements and the action ($\tilde{a}$) ones. The distinction between active and sensory states is dictated by the definition of a Markov blanket; namely, active states influence but are not influenced by external system elements, while sensory states influence but are not influenced by the representational system.

\vspace{3mm} 

In the notions offered by Friston, both the external system ($\tilde{\psi}$ units) and the representational system ($\tilde{r}$ units) each independently come to a free energy minimum. The activation of $\tilde{r}$ units within the representational system can be mediated by certain parameters ($\theta$), so that the representational system models the external one. 

More appropriately, since we are specifying that both the external and representational systems come to free energy minima, we would state that the nature of the representational system, when it achieves free energy minimization, approximates that of the external system, which also comes to free energy minimization. The degree-of-closeness of the model approximation to the external system is mediated by the parameter(s) $\theta$. 

In order to create such a computational engine, we need a formalism that will actually allow this free energy minimization to take place, in both the external and representational components of the system. Maren \cite{Maren_2016_CVM-primer-neurosci} has suggested that a free energy formulation, known as the cluster variation method (CVM), can potentially serve in such a computational engine, as is shown in Fig.~\ref{fig:CVM-2-D_comput-engine_crppd_2017-05-13}. We develop this further in Section~\ref{sec:var-FE-new-comput-engine}.

As a first step, we will use Friston's framework for the variational Bayes approach, and this requires that we derive the basic variational Bayes equations. 

To do this, we use the same notation as used by Friston, presented in the following Table~\ref{tbl:var-free-energy-eqn-variables}. The tilde notation (with variables $\tilde{\psi}$, $\tilde{r}$, $\tilde{s}$, and $\tilde{a}$) all refer to these as being ``generalized'' variables \cite{Friston-et-al_2015_Knowing-ones-place-free-energy-pattern-recognition}.

\begin{table}[ht!]\footnotesize
\label{tbl:var-free-energy-eqn-variables}
     \centering 
     \caption{Variable definitions for variational free energy equations}
     \begin{tabular}{|p{4cm}|p{7cm}|}  
        	
     \hline
	 \multicolumn{1}{|>{\centering\arraybackslash}m{4cm}|}	{\textbf{Variable}} 
    & \multicolumn{1}{>{\centering\arraybackslash}m{7cm}|}{\textbf{Meaning}} \T\B \\ 
		\hline
  
	 \multicolumn{1}{|>{\centering\arraybackslash}m{4cm}|}	{$\tilde{s},\tilde{a},\tilde{r}$} 
    & Generalized expressions for sensory ($\tilde{s}$), active ($\tilde{a}$), and internal (or representational) ($\tilde{r}$) states  \\ [5pt]   

	 \multicolumn{1}{|>{\centering\arraybackslash}m{4cm}|}	{$\tilde{\psi}$} 
    & States of the world (system being modeled) that cause sensory states, and which can be influenced by action  \\ [5pt]

	 \multicolumn{1}{|>{\centering\arraybackslash}m{4cm}|}	{$f_x(\tilde{\psi}, \tilde{s},\tilde{a},\tilde{r})$} 
    & Flow of system's states, where $x$ corresponds to external, internal (representational), or active states; see, e.g., Friston et al. (2015) \cite{Friston-et-al_2015_Knowing-ones-place-free-energy-pattern-recognition}   \\ [10pt]   
  
     \hline
     \end{tabular}
\end{table}

%
\section{The Variational Free Energy }
\label{sec:var-free-energy}
%

The goal of this section is to follow the Friston approach and express variational free energy as an expected energy or \textit{enthalpy} minus the entropy of a variational (i.e., approximate posterior) probability density. This can be equivalently expressed as \textit{surprisal} plus the reverse Kullback-Leibler (KL) divergence between the ``variational density and the posterior density over external states'' (p. 4, \cite{Friston-et-al_2015_Knowing-ones-place-free-energy-pattern-recognition}), as shown in Eqn.~\ref{eqn:var-free-energy-eqn_part2-first-time}. (For a discussion of the reverse K-L divergence, see Maren (2024) \cite{Maren_2024_Minding-Your-Ps-and-Qs-Kullback-Leibler-Divergence}.) 

(\textbf{\textit{Note:}} Most of us would say that the formulation of Eqn.~\ref{eqn:var-free-energy-eqn_part2-first-time} is between the true posterior density over external states $q({\tilde{{\psi}}}|\tilde{r})$ (given the internal or representational states and their Markov blanket) and the variational density $p(\tilde{{\psi}}|\tilde{s},\tilde{a},\tilde{r})$ (the density of the model system).)

Friston expresses the variational free energy of an ensemble as the following equations \cite{Friston_2013_Life-as-we-know-it, Friston-et-al_2015_Knowing-ones-place-free-energy-pattern-recognition} (where the exact notation is taken from Friston (2015) \cite{Friston-et-al_2015_Knowing-ones-place-free-energy-pattern-recognition}, Eqn. 3.2)

%
%

\begin{equation}
\label{eqn:var-free-energy-eqn_part1}
  \begin{aligned}
  f_r(\tilde{s},\tilde{a},\tilde{r}) = (Q_r - \Gamma_r) \nabla_{\tilde{r}}F  \\
  f_a(\tilde{s},\tilde{a},\tilde{r}) = (Q_a - \Gamma_a) \nabla_{\tilde{a}}F 
    \end{aligned} 
\end{equation}

\noindent and

\begin{equation}
\label{eqn:var-free-energy-eqn_part2-first-time}
  \begin{aligned}
  F(\tilde{s},\tilde{a},\tilde{r}) =
    E_q[L(\tilde{x})]
    - H[q(\tilde{{\psi}}|\tilde{r})] \\
    = L(\tilde{s},\tilde{a},\tilde{r}) + 
    D_{KL}[q({\tilde{{\psi}}}|\tilde{r})||
    p(\tilde{{\psi}}|\tilde{s},\tilde{a},\tilde{r})].    
  \end{aligned}
\end{equation}

\noindent where the variables were previously identified in  Table~\ref{tbl:var-free-energy-eqn-variables}. 

The flow of system states, represented by Eqn.~\ref{eqn:var-free-energy-eqn_part1},  is particular to each type of unit, so that $f_a(\tilde{s},\tilde{a},\tilde{r})$ is the (gradient descent) change in the set of action units ($\tilde{a}$), and $f_s(\tilde{s},\tilde{a},\tilde{r})$ is the change in the set of action units ($\tilde{s}$). These system states are subject to random fluctuations denoted by $\omega$. The amplitude of the random fluctuations is controlled by [the diffusion tensor] $\Gamma$, while (the set of) $Q$ are antisymmetric matrices that allow for solenoidal flow (which does not change free energy). (See further discussion in Sect. 2 of Friston (2013) \cite{Friston_2013_Life-as-we-know-it}.)

This Technical Report focuses exclusively on the static Eqn.~\ref{eqn:var-free-energy-eqn_part2-first-time}, and defers the dynamic Eqn.~\ref{eqn:var-free-energy-eqn_part1} to a different occasion. 

Eqn.~\ref{eqn:var-free-energy-eqn_part2-first-time} expresses the \textit{variational free energy} $F(\tilde{s},\tilde{a},\tilde{r})$, which is isomorphic in structure to the \textit{free energy} used in statistical thermodynamics. In Eqn.~\ref{eqn:var-free-energy-eqn_part2-first-time}, this term is expressed in two ways. First, it given as the difference of the expectation of $L  (\tilde{x})$ (a log-likelihood term) and the  \textit{entropy of the posterior density over external states}, $H[q]$. (This is the formalism that is isomorphic with statistical thermodynamics.) We will determine the exact meaning of $L (\tilde{x})$ in Section~\ref{sec:var-FE-log-likelihood-expectation-and-entropy}.

The second expression for $F(\tilde{s},\tilde{a},\tilde{r})$ is the sum of the negative log evidence (\textit{surprisal}) $L(\tilde{s},\tilde{a},\tilde{r})$ and the \textit{reverse} Kullback-Leibler (K-L) divergence between the external system and the model. We will determine the meaning of $L(\tilde{s},\tilde{a},\tilde{r})$ in Section \ref{sec:var-FE-log-likelihood-and-KL-divergence}, and identify how (and why) it is different from the $L(\tilde{x})$ used in the first expression. (As a minor note: as the reverse Kullback-Leibler divergence in this expression approaches zero, we see that the variational free energy can be identified as an \textit{evidence bound} for the system.)

We further note that a primary difference between these two expressions is that in the first expression, $H[q(\tilde{{\psi}}|\tilde{r})]$ is an exact entropy term, whereas in the second expression, the $D_{KL}$ term is a \textit{relative} entropy. 

The following Table~\ref{tbl:glossary-thermodynamic} presents a glossary of the thermodynamic terms used in this Report.

%
\begin{table}[ht!]\footnotesize
	\label{tbl:glossary-thermodynamic}
     \centering 
     \caption{Variable definitions for variational free energy equations}
     \vspace{3mm}
     \begin{tabular}{|p{3cm}|p{10cm}|}  
        	
     \hline
	 \multicolumn{1}{|>{\centering\arraybackslash}m{3cm}|}	{\textbf{Variable}} 
    & \multicolumn{1}{>{\centering\arraybackslash}m{10cm}|}{\textbf{Meaning}} \T\B \\ 
		\hline
		
	 \multicolumn{1}{|>{\centering\arraybackslash}m{3cm}|}	{Activation enthalpy} 
    &Enthalpy  $\varepsilon_0$ associated with a single unit (node) in the ``on'' or ``active'' state (\textbf{A}); influences configuration variables and is set to 0 in order to achieve an analytic solution for the free energy equilibrium  \\ [3pt] 		

	 \multicolumn{1}{|>{\centering\arraybackslash}m{3cm}|}	{Configuration variable(s)} 
    &Nearest neighbor, next-nearest neighbor, and triplet patterns  \\ [8pt] 

	 \multicolumn{1}{|>{\centering\arraybackslash}m{3cm}|}	{Degeneracy} 
    &Number of ways in which a configuration variable can appear  \\ [5pt] 

	 \multicolumn{1}{|>{\centering\arraybackslash}m{3cm}|}	{Enthalpy} 
    &Internal energy \textit{H} results from both per unit and pairwise interactions; often denoted $H$ in thermodynamic treatments    \\ [5pt]  
    
	 \multicolumn{1}{|>{\centering\arraybackslash}m{3cm}|}	{Entropy} 
    &The entropy \textit{S} is the distribution over all possible states; often denoted $S$ in thermodynamic treatments and $H$ in information theory \\ [5pt]       

	 \multicolumn{1}{|>{\centering\arraybackslash}m{3cm}|}	{Equilibrium point} 
    &By definition, the free energy minimum for a closed system  \\ [5pt] 
 
	 \multicolumn{1}{|>{\centering\arraybackslash}m{3cm}|}	{Equilibrium distribution} 
    & Configuration variable values when free energy minimized for given \textit{h}  \\ [5pt]   

	 \multicolumn{1}{|>{\centering\arraybackslash}m{3cm}|}	{Ergodic distribution} 
    & Achieved when a system is allowed to evolve over a long period of time   \\ [5pt]  
 
	 \multicolumn{1}{|>{\centering\arraybackslash}m{3cm}|}	{Free Energy} 
    &The thermodynamic state function \textit{F}; where \textit{F = H-TS}; sometimes \textit{G} is used instead of \textit{F}; referring to (thermodynamic) Gibbs free energy   \\ [5pt]       
    
	 \multicolumn{1}{|>{\centering\arraybackslash}m{3cm}|}	{\textit{h-value}} 
    & A more useful expression for the interaction enthalpy parameter $\varepsilon_1$; $h = e^{2\beta\varepsilon_1}$, where $\beta = 1/{k_{\beta}T}$,  and where $k_{\beta}$ is Boltzmann's constant and $T$ is temperature; $\beta$ can be set to 1 for our purposes   \\ [5pt]       

	 \multicolumn{1}{|>{\centering\arraybackslash}m{3cm}|}	{Interaction enthalpy} 
    & Between two unlike units, $\varepsilon_1$; influences configuration variables  \\ [7pt] 
 
	 \multicolumn{1}{|>{\centering\arraybackslash}m{3cm}|}	{Interaction enthalpy parameter} 
    & Another term for the \textit{h-value} where $h=e^{2\varepsilon_1}$  \\ [9pt]   
    
	 \multicolumn{1}{|>{\centering\arraybackslash}m{3cm}|}	{Temperature} 
    &Temperature \textit{T} times Boltzmann's constant $k_{\beta}$ is set equal to one \\ [10pt]    
	  
     \hline
     \end{tabular}
\end{table}
%

%
\subsection{The Final Result in a Nutshell }
\label{subsec:final-results}
%

The essence of Eqn.~\ref{eqn:var-free-energy-eqn_part2-first-time} is that we are taking a single expression, and parsing and re-organizing it to achieve two different ways of re-expressing the same thing. The expression, sometimes called the ``variational free energy'' (see, e.g., Friston (op. cit.)) is given as

\begin{equation}
\label{eqn:Friston-free-energy-integral-first-time}
  \begin{aligned}
    F(\tilde{s},\tilde{a},\tilde{r})
     = - \int_{\psi} q({\tilde{{\psi}}}|\tilde{r}) \ln\left({\frac 
    {p(\tilde{{\psi}},\tilde{s},\tilde{a},\tilde{r})}
     {q({\tilde{{\psi}}}|\tilde{r})}}
    \right)d{\psi}.  
  \end{aligned}
\end{equation}

\begin{figure}[htbp]
    \centering
        \includegraphics[trim=0cm 0cm 0cm 0cm, clip=true,  width=0.90\textwidth]{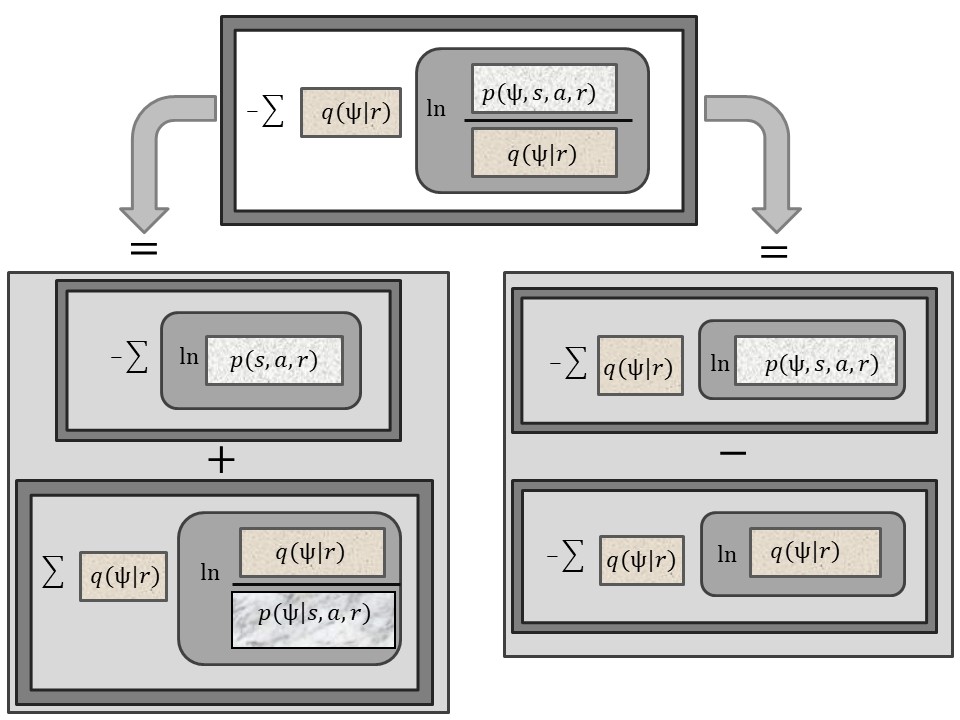}
   \caption{Diagrammatic illustration of Eqn.~\ref{eqn:var-free-energy-eqn_part2-first-time}.} 
    \label{fig:var-free-energy-eqn_part2-first-time}
\end{figure}

Diagrammatically, we can see this shown in Figure~\ref{fig:var-free-energy-eqn_part2-first-time}. The tilde notation is dropped in this figure and in the immediately-following subsection, which discusses this figure.

%
\subsection{Quick Summary of Key Points }
\label{subsec:quick-summary}
%

In Figure~\ref{fig:var-free-energy-eqn_part2-first-time}, we saw that the initiating expression, the ``variational free energy,'' was being reorganized in two different ways. On the Left-Hand-Side (LHS), we saw that it led to the (negative of) a simple sum over the log-likelihood of the set of variables associated with the representational units and the Markov blanket, added to the reverse Kullback-Leibler (K-L) divergence term expressing the difference between the model $q$ and the probability distribution of the external units $\psi$ \textit{as conditioned on} the internal units and the Markov blanket. We achieve this result by re-expressing the probability of the joint co-occurrence, $p(\psi, s,a,r)$ as a conditional probability, using the formulation for a Bayesian posterior.

On the Right-Hand-Side (RHS), we see that the re-organization is much simpler, and can indeed be followed simply by examining the diagram itself. (There are a few subtleties, which are addressed in the following sections.) The first term on the RHS is a weighted sum over the joint probability distribution $p(\psi, s,a,r)$. The second term is a term that looks remarkably like entropy. 

The structure of the equation on the RHS, and the ``entropy-like'' appearance of the last term, has given rise to expressing the whole equation as ``variational free energy.'' This is because there is a morphological similarity between the \textit{form} of the variational free energy equation and the classic thermodynamic free energy. (See Appendix~\ref{sec:Fund-thermo-concepts} for a quick review of the basic statistical thermodynamics equations, with  more details presented in Appendix~\ref{sec:Appendix-var-FE-enthalpy-entropy}.)

%
\subsection{Various Iinterpretations }
\label{subsec:various-interpretations}
%

Various authors interpret Eqn.~\ref{eqn:var-free-energy-eqn_part2-first-time} with different notations and descriptive phrases. The purpose of this subsection is to identify a few of these interpretations, and to tease out exactly what is meant from exactly what is said. This should make it easier for those reading the source papers to understand what is actually being presented, and is a first step towards building a \textit{Rosetta stone}; giving a cross-correlation between two different notations.

The specifics of this \textit{Rosetta stone} are captured in Table~\ref{tbl:Beal-Friston-Blei-notation}, presented in Section~\ref{sec:var-FE-log-likelihood-and-KL-divergence}.

For example, Friston (2013) presents this Report's Eqn.~\ref{eqn:var-free-energy-eqn_part2-first-time} as his Eqn. 2.7 in \cite{Friston_2013_Life-as-we-know-it}, using the notation

\begin{equation}
\label{eqn:var-free-energy-eqn_part2-different-notation}
  \begin{aligned}
  F(s,a,\lambda) =
    E_q[G(\psi,s,a,\lambda)]
    - H[q({\psi}|\mu)]. \nonumber    
  \end{aligned}
\end{equation} 

Friston refers to $E_q[G]$ saying that ``[T]he last equality just shows that free energy can be expressed as the expected Gibbs energy minus the entropy of the variational density.'' 

However, Sengupta, Stemmler, and Friston \cite{Sungupta-et-al_2013_Info-efficiency-nervous-system} state that ``$U(t) = - \ln p(s(t), \psi(t)|m)$ corresponds to an internal energy under a generative model of the world, described in terms of the density over sensory and hidden states $p(s,y|m)$.'' (\textit{Author's note:} $U(t)$ corresponds to $E_q[G]$, from \cite{Sungupta-et-al_2013_Info-efficiency-nervous-system} and \cite{Friston_2013_Life-as-we-know-it}, respectively.)  Moreover, Sengupta et al. state that ``$F(t)$ is called free energy —- by analogy with its thermodynamic homologue that is defined as internal energy minus entropy. However, it is important to note that variational free energy is not the Helmholtz free energy ... it is a functional of a probability distribution over hidden (fictive) states encoded by internal states $q(y|m)$, not the probability distribution over the (physical) internal states. This is why variational free energy pertains to information about hidden states that are represented, not the internal states that represent them.'' 

(\textit{Author's Note 1:} For the benefit of those who wish to compare the information-theoretic approach of Beal, Friston, and others against a classic statistical thermodynamics formulation, Appendix~\ref{sec:Fund-thermo-concepts} derives fundamental thermodynamic concepts, and Beal's results compared with the corresponding statistical thermodynamic formalism are given in Appendix B. As stated by Sengupta et al., they are not precisely the same \cite{Sungupta-et-al_2013_Info-efficiency-nervous-system}.) 

(\textit{Author's Note 2:} Sengupta et al. \cite{Sungupta-et-al_2013_Info-efficiency-nervous-system} refer to a \textit{Helmholtz} free energy, and Friston, writing separately, refers to a \textit{Gibbs} free energy. Both Helmholtz and Gibbs free energies correspond to thermodynamic free energy formulations, and involve measurements on a physical system, i.e., temperature, pressure, and volume. The distinction between Helmholtz and Gibbs free energies disappears when we use the thermodynamic free energy formulations as a metaphor.)

It will be clear, in the succeeding derivations, that what is offered as $E_q[G]$ is not what we are familiar with as the Gibbs (or Helmholtz) free energy from statistical thermodynamics. However, Sengupta et al. offer the following explanation as a \textit{Lemma}:

``\textit{\textbf{Lemma}: (complexity minimisation)} Minimising the complexity of a conditional distribution —- whose sufficient statistics are (strictly increasing functions of) some unconstrained internal variables of a thermodynamic system —- minimises the Helmholtz free energy of that system.''

As proof, they suggest that we can use standard results from Bayesian statistics  \cite{Beal_2003_Variational-algorithm-approx-Bayes-inference} in order to express free energy as complexity minus accuracy. They are, in fact, referring to the derivation in Beal (2003) that will be the fundamental reference in this Technical Report. They conclude that  ``In sum, the internal states encoding prior beliefs about hidden states of the world are those that minimise Helmholtz free energy and the complexity defined by variational free energy.''

Friston cites Beal (2003) \cite{Beal_2003_Variational-algorithm-approx-Bayes-inference} for the derivation of Eqns.~\ref{eqn:var-free-energy-eqn_part1} and \ref{eqn:var-free-energy-eqn_part2-first-time}. Blei et al. \cite{Blei-et-al_2016_Variational-Bayes} also provide a useful tutorial. The following sections walk through the derivations as provided by Beal, using some of the material provided by Blei et al. to support and elucidate certain points. The goal is to make the match between the free energy equations as expressed by Beal (and in certain cases, by Blei et al.) with those used by Friston as clear and as transparent as possible.

%
\section{Important Distinction \& Clarification }
\label{sec:important-distinction-and-clarification}
%

Following the transition of the variational Bayes approach from Beal \cite{Beal_2003_Variational-algorithm-approx-Bayes-inference} to Friston (op. cit.), and from thence to an actual, computable model, requires some subtlety and attention to detail. The most crucial consideration at the outset is that with Beal (and with previous expostulations on the variational Bayes method), both the actual data points being modeled and the model itself are expressed with regard to some underlying variables, $x_i$. Thus, we have the external (or observable, or dependent) variables $y_i = y(x_i)$, and the distribution $p_{x_i}(x_i)$.

In contrast to Beal's description, with Friston, the external system and the representation are separated by a Markov blanket. Thus, the external system's units, denoted $\psi$, are distinct from the representational system's units. Friston's approach is illustrated in Figure~\ref{fig:Fristons-notation}.

({\textbf{\textit{Note:}} Beal actually does address how variational inference is performed in a system separated by a Markov blanket; see Sections 3 ff. of his work; his notation gets complex, and we don't need to use his more detailed work in order to interpret Friston.) 

\begin{figure}[htbp]
    \centering
        \includegraphics[trim=0cm 0cm 0cm 0cm, clip=true,  width=0.80\textwidth]{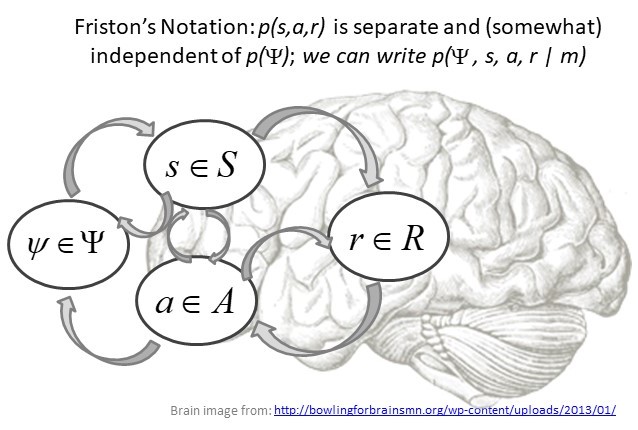}
   \caption{In the variational Bayes method described by Friston, the external system, whose units are denoted by $\psi$, interacts with a separate representational system whose units are denoted by $r$. The two systems are separated by a Markov blanket composed of sensing ($s$) and action ($a$) units. (\textit{Note:} for simplicity, the tilde notation is dropped from this figure.) In comparison with Beal, the distribution $q$ is of the external system, which is conditioned on the representational system; $q = q(\tilde{\psi}|\tilde{r})$. This is feasible because the external system units $\tilde{\psi}$ influence the representational units $\tilde{r}$ through the sensory units $\tilde{s}$. Conversely, the representational units $\tilde{r}$ influence the external units $\tilde{\psi}$ through the active units $\tilde{a}$. } 
    \label{fig:Fristons-notation}
\end{figure}

In a beautiful and intriguing illustration, Friston applies his formulation to the emergence of a Markov blanket, beginning with ``an ensemble of elemental subsystems with (heuristically speaking) Newtonian and electrochemical dynamics ... One can think of these generalized states as describing the physical and electrochemical state of large macromolecules. Crucially, these states are coupled within and between the subsystems comprising an ensemble'' \cite{Friston_2013_Life-as-we-know-it}. The units comprising the Markov blanket and the internal, ``representational'' system emerge over time. Appropriately enough, Friston's illustration focuses on the dynamic behaviors of the various units.

The following Table~\ref{tbl:glossary-information-theoretic} presents a glossary of the thermodynamic terms used in this Report.

%
\begin{table}[ht!]\footnotesize
	\label{tbl:glossary-information-theoretic}
     \centering 
     \caption{Variable definitions for information theoretic terms}
     \vspace{3mm}
     \begin{tabular}{|p{3cm}|p{10cm}|}  
        	
     \hline
	 \multicolumn{1}{|>{\centering\arraybackslash}m{3cm}|}	{\textbf{Variable}} 
    & \multicolumn{1}{>{\centering\arraybackslash}m{10cm}|}{\textbf{Meaning}} \T\B \\ 
		\hline

	 \multicolumn{1}{|>{\centering\arraybackslash}m{3cm}|}	{Kullback-Leibler divergence} 
    &Relative entropy, or measure of how one probability distribution differs from a second one; here, the divergence between a model distribution and the actual system being modeled \\ [3pt] 	
		
	 \multicolumn{1}{|>{\centering\arraybackslash}m{3cm}|}	{Surprisal} 
    &Information content, or amount of information gained when a random variable is sampled  \\ [3pt] 		

	 \multicolumn{1}{|>{\centering\arraybackslash}m{3cm}|}	{Variational free energy $F(\tilde{s},\tilde{a},\tilde{r})$} 
    &Isomorphic in structure to the equation for thermodynamic free energy, this expresses the difference between the expectation for the probability distribution over a system and the entropy of a model system; it can also be expressed as the sum of the negative log evidence (surprisal) and the Kullback-Leibler divergence between the model and the external system  \\ [3pt]

     \hline
     \end{tabular}
\end{table}
%

%
\subsection{Primary Distinction: Beal and Friston}
\label{subsec:primary-distinction}
%

Turning our attention back to the evolution of Friston's formulation from that expressed by Beal, we investigate the correspondence between Beal's expression of $p(y_i|\theta)$ in comparison with Friston's use of $p(\tilde{s},\tilde{a},\tilde{r})$. 

In writing $p(y_i|\theta)$, Beal is actually describing the observable variables that are being modeled. He is specifically referring to the (integration of) the joint probability distribution $p(x_i, y_i|\theta)$, where the $y_i$ are the dependent, or observable variables, and the $x_i$ are the independent, or hidden, or latent variables. Specifically, Beal states (Sect. 2.2.1 \cite{Beal_2003_Variational-algorithm-approx-Bayes-inference}), ``Consider a model with hidden variables $x$ and observed variables $y$. The parameters describing the (potentially) stochastic dependencies between variables are given by $\theta$.'' 

The key to understanding that Beal is describing the external (observable) system (also that which Friston refers to as being denoted by the units $\tilde{\psi}$) is that both Beal and Friston refer to the \textit{distribution of hidden variables} as $p(x_i)$ (see Eqn.~\ref{eqn:var-free-energy-eqn_part2-first-time} as an example). As Friston notes (personal communication), ``The link between the two formulations rests upon associating the causes $x_i$ and $\theta$ in Beal's notation with the external states that cause data ($y_i$) in Friston's formulation (i.e., $x_i$ and $\theta_i$ correspond to $\tilde{\psi}$), where the data in Friston's formulation become the sensory units states ($\tilde{s}$).''

In support of this, we note that Blei et al. \cite{Blei-et-al_2016_Variational-Bayes} stated that ``the approach  is to posit a family of approximate densities \textbf{\textit{Q}}, which are defined over the set of latent variables $z = z_{1:m}$ and observations $x = x_{1:n}$ ... Then, we try to find that member \textit{q*(z)} of the set \textbf{\textit{Q}} that is the Kullback-Leibler divergence (KL) of \textit{q*(z)} with respect to the exact posterior $p(z|x)$, which represents the probability distribution of the latent variables with regard to the observables, given as

\begin{equation}
\label{eqn:argmin-for-KL-restated}
  \begin{aligned}
    \textit{q*(\textbf{z})} = arg min KL(\textit{q(\textbf{z})}||(p(z|x)); q(z)\in \textbf{\textit{Q}}.''  \nonumber  
  \end{aligned}
\end{equation}

Note that in this explanation by Blei et al., the observable variable was denoted as $x$ instead of $y$, but the independent (and hidden) variable was denoted $z$. Thus, in Beal, the observables are $y$ and the latent are $x$, and in Blei et al., the observables are $x$ and the latent are $z$. (For a useful cross-correlation of notation across all three authors, see Table~\ref{tbl:Beal-Friston-Blei-notation}.)

The important thing to note, and the thing that differentiates Friston's formulation from Beal's, is that with Beal, both $p$ and $q$ are applied to the same underlying hidden or independent variable(s) $x$. Essentially, $p$ represents the exact observed data, and $q$ is the model (subject to parameter fitting with $\theta$). In contrast, with Friston, we are dealing with entirely different sets of units; the external units (denoted by $\tilde{\psi}$) and the internal, representational units (denoted by $\tilde{r}$). We can express a probability distribution over each, but they are not necessarily the respective observed data points and modeling functions taken over the same underlying base set of values.

%
\subsection{Integrating over the Model Space}
\label{subsec:intgrating-over-model}
%

As we will note during subsequent derivations (presented in Subsection~\ref{subsec:interpreting-KL-divergence}), the integration (in various equations) over $q$ is shown as being with regard to the external system units, $\tilde{\psi}$. More to the point, Friston uses the notation of integrating over $\tilde{\psi}$, in order to show the comparison between his formulation and that of Beal. 

Specifically, we will see that Beal uses the integration (Section~\ref{sec:var-FE-log-likelihood-expectation-and-entropy})

\begin{equation}
\label{eqn:Beal-free-energy-integral-early-illustration}
  \begin{aligned}
    L(\theta)  \geq  \sum_{i=1} \int dx_i \: q_{x_i}(x_i) \: \ln \: \frac { p(x_i,y_i|\theta)} {q_{x_i}{(x_i})}, \nonumber  
  \end{aligned}
\end{equation}

\noindent
and correspondingly, Friston uses the integration (Subsection~\ref{subsec:rewriting-KL-divergence})

\begin{equation}
\label{eqn:Friston-free-energy-integral-early-illustration}
  \begin{aligned}
    F(\tilde{s},\tilde{a},\tilde{r})
     = - \int_{\psi} q({\tilde{{\psi}}}|\tilde{r}) \ln\left({\frac 
    {p(\tilde{{\psi}},\tilde{s},\tilde{a},\tilde{r})}
     {q({\tilde{{\psi}}}|\tilde{r})}}
    \right)d{\psi}. \nonumber    
  \end{aligned}
\end{equation}

Friston is clearly making an effort to show the correspondence between his formulation and that of Beal. Also, as he envisions it, a summation or integration of the model over the external system may be possible. (His illustration in \cite{Friston_2013_Life-as-we-know-it} showed the evolution of the internal states and the Markov blanket from an original ``primordial soup'' encompassing all the units.)

For the purposes of this Technical Report, though, we envision a system where the distribution $q$ refers strictly to the external units $\tilde{\Psi}$ as conditioned by the internal (representational) units $\tilde{r}$. Over time, the goal is to adjust the units $\tilde{r}$ so that the free energy of the distribution $q$ approximates that of the actual external system with units $\tilde{\psi}$. 

Thus, the elements of our system that we've been considering so far consist of three things: 
\begin{enumerate}
\item The external system which is composed of units $\tilde{\psi}$; we are trying to model this, and we operate under the presumption that we cannot always directly compute certain measures on this system, 
\item The internal system which is composed of units $\tilde{r}$; at any given moment we can determine certain measures on this system, yielding $L(\tilde{s},\tilde{a},\tilde{r})$  (we are temporarily ignoring $\tilde{s}$ and $\tilde{a}$), and 
\item A distribution of the external system expressed via the internal system, $q$, where the chief distinction is that when we take an actual value for $q$, we do so with the presumption that the internal system is brought to a free energy equilibrium for a given set of parameter values $\theta$. This means that the measures for a given distribution-in-the-moment, as represented by $L$, would be adjusted to represent what they \textit{would} be if the internal system were brought to equilibrium, for a specific set of $\theta$.  
\end{enumerate}

Thus, when it comes to the integration steps, we will consider that an integration of the distribution $q$ over $\tilde{\psi}$ will be interpreted as integrating over the distribution units themselves ($\tilde{r}$), but with consideration that the distribution $p$ will have come into a free energy equilibrium (subject to parameters $\theta$) that is a best approximation, for that set of $\theta$, to the external system.

%
\section{The Variational Free Energy: Reverse K-L divergence and Log-Llikelihood }
\label{sec:var-FE-log-likelihood-and-KL-divergence}
%

In Eqn.~\ref{eqn:var-free-energy-eqn_part2-first-time}, the free energy is expressed in two different ways: 

\begin{enumerate}
\item As the difference between an enthalpy-like term and the entropy of the distribution $q$, and
\item As the sum of a surprisal or potential (negative log evidence) and the (reverse) K-L divergence between the probability distributions of the model and the  external system. 
\end{enumerate}

The following Table~\ref{tbl:Beal-Friston-Blei-notation} presents a ``Rosetta Stone'' of the differing notations as used by Beal, Friston, and Blei et al. 

Notation regarding use of a Markov blanket is presented for Friston only; while Beal does address Markov blankets in his work, we use here only the simpler form of his notation, and similarly use only simple notation presented by Blei et al.

%
\begin{table}[ht!]\footnotesize
	\label{tbl:Beal-Friston-Blei-notation}
     \centering 
     \caption{The \textbf{\textit{Rosetta Stone}}: Beal, Friston, and Blei Notation}
     \vspace{3mm}
     \begin{tabular}{|p{5cm}|p{2cm}|p{2cm} |p{2cm}|}  
        	
     \hline
	 \multicolumn{1}{|>{\centering\arraybackslash}m{5cm}|}	{\textbf{Variable / Notation}} 
    & \multicolumn{1}{>{\centering\arraybackslash}m{2cm}|}{\textbf{Beal}} 
    & \multicolumn{1}{>{\centering\arraybackslash}m{2cm}|}{\textbf{Friston}}     
    & \multicolumn{1}{>{\centering\arraybackslash}m{2cm}|}{\textbf{Blei}} \T\B  \\ 
		\hline

	 \multicolumn{1}{|>{\centering\arraybackslash}m{5cm}|}	{\textbf{Observable Variable}; \textit{ Dependent or ``Internal States'}'}
    & \multicolumn{1}{>{\centering\arraybackslash}m{2cm}|}{$y_i$}
    & \multicolumn{1}{>{\centering\arraybackslash}m{2cm}|}{$\lambda, \tilde{r}$}
    & \multicolumn{1}{>{\centering\arraybackslash}m{2cm}|}{$x_i$}   \\  [16pt] 	
    
	 \multicolumn{1}{|>{\centering\arraybackslash}m{5cm}|}	{\textbf{Hidden Variable}; \textit{ Independent, Latent, or ``External States''}}
    & \multicolumn{1}{>{\centering\arraybackslash}m{2cm}|}{$x_i$}
    & \multicolumn{1}{>{\centering\arraybackslash}m{2cm}|}{$\tilde{\Psi}$}
    & \multicolumn{1}{>{\centering\arraybackslash}m{2cm}|}{$z_i$}   \\  [16pt] 	    
    
	 \multicolumn{1}{|>{\centering\arraybackslash}m{5cm}|}	{\textbf{Markov ``sensing'' units}}
    & \multicolumn{1}{>{\centering\arraybackslash}m{2cm}|}{ - }
    & \multicolumn{1}{>{\centering\arraybackslash}m{2cm}|}{$\tilde{s}$}
    & \multicolumn{1}{>{\centering\arraybackslash}m{2cm}|}{ - }   \\  [8pt] 	
    
	 \multicolumn{1}{|>{\centering\arraybackslash}m{5cm}|}	{\textbf{Markov ``active'' units}}
    & \multicolumn{1}{>{\centering\arraybackslash}m{2cm}|}{ - }
    & \multicolumn{1}{>{\centering\arraybackslash}m{2cm}|}{$\tilde{a}$}
    & \multicolumn{1}{>{\centering\arraybackslash}m{2cm}|}{ - }   \\  [8pt] 	     
    
	 \multicolumn{1}{|>{\centering\arraybackslash}m{5cm}|}	{\textbf{Model parameters}}
    & \multicolumn{1}{>{\centering\arraybackslash}m{2cm}|}{$\theta$ }
    & \multicolumn{1}{>{\centering\arraybackslash}m{2cm}|}{$m$}
    & \multicolumn{1}{>{\centering\arraybackslash}m{2cm}|}{ - }   \\  [8pt] 

	 \multicolumn{1}{|>{\centering\arraybackslash}m{5cm}|}	{\textbf{Model distribution}}
    & \multicolumn{1}{>{\centering\arraybackslash}m{2cm}|}{$q(x)$ (1)}
    & \multicolumn{1}{>{\centering\arraybackslash}m{2cm}|}{$q(\Psi|\lambda)$ (2)}
    & \multicolumn{1}{>{\centering\arraybackslash}m{2cm}|}{ - }   \\  [8pt]     
    
	 \multicolumn{1}{|>{\centering\arraybackslash}m{5cm}|}	{\textbf{Observations distribution}}
    & \multicolumn{1}{>{\centering\arraybackslash}m{2cm}|}{$p(y|\theta)$ (3)}
    & \multicolumn{1}{>{\centering\arraybackslash}m{2cm}|}{$p(\Psi, s, a, r| m)$ (4)}
    & \multicolumn{1}{>{\centering\arraybackslash}m{2cm}|}{ - }   \\  [16pt]     
    		
	 \multicolumn{1}{|>{\centering\arraybackslash}m{5cm}|}	{\textbf{Variational free energy}}
    & \multicolumn{1}{>{\centering\arraybackslash}m{2cm}|}{ - }
    & \multicolumn{1}{>{\centering\arraybackslash}m{2cm}|}{$F(\tilde{s},\tilde{a},\tilde{r})$}
    & \multicolumn{1}{>{\centering\arraybackslash}m{2cm}|}{ - }   \\  [8pt] 		


     \hline
     \end{tabular}
\end{table}
%

The authors specifically identify their notation, according to the following enumerated points (corresponding to elements of Table~\ref{tbl:Beal-Friston-Blei-notation}):

\begin{enumerate}
\item \textbf{Model distribution - Beal:}  $q_{x_i}(x_i)$: ``we use a distinct distribution $q_{x_i}(x_i)$ over the hidden variables ...''  (Beal, 2003,  p. 47, just before Eqn. 2.12),
\item \textbf{Model distribution - Friston:} $q(\Psi|\lambda)$: `` ...  a probability density over external states $q(\Psi|\lambda)$ that is encoded (parametrized) by internal states.''  (Friston, 2013, p. 4, just before Lemma 2.1).
\item \textbf{Observations - Beal:} $p(y|\theta)$: `` ... [the] generative model that produces a dataset $y = \{y_1, . . . , y_n\}$ consisting of \textit{n} independent and identically distributed (i.i.d.) items, generated using a set of hidden variables $x = \{x_1, . . . , x_n\}$ such that the likelihood can be written as a function of $\theta$ ...'' (Beal, 2003, p.46, Eqn. 2.9), and
\item \textbf{Observations - Friston:} $p(\Psi, s, a, r| m)$: ``...  ergodic density $p(\Psi, s, a, r| m)$ [is] a probability density function over external $\psi \in \Psi$, sensory $s \in S$, active $a \in A$ and internal states $\lambda \in \Lambda$  for a system denoted by $m$'' (Friston, 2013, p. 2, Table 1). 
\end{enumerate}

In this section, we focus on the second half of Eqn.~\ref{eqn:var-free-energy-eqn_part2-first-time}; the equality between the ``variational free energy'' and the sum of the pooled negative log probabilities of sensory states (and their accompanying representational and active states) and the reverse K-L divergence. Specifically, we wish to show that

\begin{equation}
\label{eqn:var-free-energy-eqn_part2-first-time-second-half}
  \begin{aligned}
  F(\tilde{s},\tilde{a},\tilde{r}) 
    = L(\tilde{s},\tilde{a},\tilde{r}) + 
    D_{KL}[q({\tilde{{\psi}}}|\tilde{r})||
    p(\tilde{{\psi}}|\tilde{s},\tilde{a},\tilde{r})].    
  \end{aligned}
\end{equation}

In achieving this goal, we will also accomplish two other tasks, namely:

\begin{enumerate}
\item Obtain a precise mathematical formation for $ F(\tilde{s},\tilde{a},\tilde{r})$, and
\item Interpret this mathematical formulation in a useful manner. 
\end{enumerate}

We begin our derivation of Eqn.~\ref{eqn:var-free-energy-eqn_part2-first-time-second-half} by first considering the definition for the reverse K-L divergence in the context of the system that we are describing (and using the notation advanced by Friston (2015) \cite{Friston-et-al_2015_Knowing-ones-place-free-energy-pattern-recognition}).

%
\subsection{Interpreting the Reverse K-L Divergence}
\label{subsec:interpreting-KL-divergence}
%

For the discrete case, we write the \textit{reverse} Kullback-Leibler (K-L) divergence as

\begin{equation}
\label{eqn:variational-K-L-divergence-summation}
  \begin{aligned}
   D_{KL}[q({\tilde{{\psi}}}|\tilde{r})||
    p(\tilde{{\psi}}|\tilde{s},\tilde{a},\tilde{r})] = \sum_{i=1}^I q({\tilde{{\psi}}}|\tilde{r}) \ln\left({\frac {q({\tilde{{\psi}}}|\tilde{r})}
    {p(\tilde{{\psi}}|\tilde{s},\tilde{a},\tilde{r})}}\right).    
  \end{aligned}
\end{equation}

For a discussion of the K-L divergence, together with the \textit{reverse} K-L divergence (which is used in all generative methods, including variational inference and active inference), see Maren (2024) \cite{Maren_2024_Minding-Your-Ps-and-Qs-Kullback-Leibler-Divergence}.

We briefly interpret the physical meaning of the terms in Eqn.~\ref{eqn:variational-K-L-divergence-summation}. The \textit{reverse} K-L divergence measures the divergence between the model-distribution $q$ of (i.e., probability distribution over) the external system, as conditioned on the representation $\tilde{r}$, against the actual distribution of the external system itself $p(\tilde{\psi}|\tilde{s},\tilde{a},\tilde{r})$. 

We note that any time we write $p(x)$, we are implicitly writing $p(x|m)$, because we are using $p$ to represent the notion of a (generative) distribution that uses a certain parameter set $\theta$.

The model-distribution $q$ is a model of the external system, $\tilde{\psi}$, which is why we write $q = q(\tilde{\psi}|\tilde{r})$. The key feature in computing $q$ is that (for the application being considered here) we take it at the equilibrium state. That is, $q$ corresponds to the \textit{equilibrium free energy} of the external system, which can be computed (or approximated) if we have a suitable free energy equation. Thus, in Eqn.~\ref{eqn:variational-K-L-divergence-summation}, we are looking at the divergence between the model-distribution of the system at equilibrium and the probabilities $p$ of various components of the system, potentially in a not-yet-at-equilibrium state. 

The parameter(s) $\theta$ directly influence $p$, but the notation for $\theta$ is suppressed in this section. 

Thus, we can read the term $q(\tilde{\psi}|\tilde{r})$ as the ``probability distribution of the model of the external system $\tilde{\psi}$, which is computed based solely on the value of the representational units $\tilde{r}$ that are isolated from the external system $\tilde{\psi}$ by a Markov blanket, but these representational units are to be considered with their at-equilibrium values.''

Next, we examine the term ${p(\tilde{{\psi}}|\tilde{s},\tilde{a},\tilde{r})}$, which expresses the probability distribution of units $\tilde{\psi}$ in the external system, conditioned on the Markov blanket sensory units $\tilde{s}$ and action units $\tilde{a}$, along with the representational units $\tilde{r}$. We recall, from the design of the entire system (external plus Markov blanket plus representational units), and also from figures given in Friston \cite{Friston_2013_Life-as-we-know-it} and Friston et al. \cite{Friston-et-al_2015_Knowing-ones-place-free-energy-pattern-recognition}, and replicated in Figure~\ref{fig:Fristons-notation}, that the representational units do not communicate directly with the external units. Thus, the dependence of $\tilde{\psi}$ is very much an implicit relationship; one that is at a distance because the direct interactions of the units in $\tilde{\psi}$ are exclusively with $\tilde{s}$ and $\tilde{a}$. Further, the system design is that the sensory units receive inputs from the external units $\tilde{\psi}$, but do not directly influence the $\tilde{\psi}$ themselves. 

Thus, the conditional relationship expressed in $p(\tilde{\psi}|\tilde{s}, \tilde{a}, \tilde{r})$ seems a little forced. However, it is the basis for our next steps in the derivation, and we will think of it simply as stating that the external system can indeed be influenced by the evolving values for the representational system $\tilde{r}$. 

With this in mind, we go back to Eqn.~\ref{eqn:variational-K-L-divergence-summation}, and interpret the \textit{reverse} K-L divergence on the Right-Hand-Side (RHS) of the equation. It states that this \textit{reverse} K-L divergence is the sum, over all possible states in which the system can possibly find itself, of the (model probability) distribution $q$ for a given specific state, multiplying the natural log of the  distribution of that state (for the actual, external system), which is divided by the actual probability for those external states. In this expression, the \textit{actual} distribution of the external system, $p(\tilde{\psi}|\tilde{s}, \tilde{a}, \tilde{r})$, is conditioned by the states of the Markov blanket and the representational system. 

Thinking ahead, we consider how the traditional formulation for how the relationship between the model and the external system is expressed, specifically as a reverse K-L divergence. Typically, the actual ``external'' system is some set of values, $p(x)$, and the model is given as $q(x)$. The (generative) distribution $p$ is specified via certain parameters $\theta$. 

In our situation, though, we will be looking at both an external system and an internal model that will each, separately, come to their respective free energy equilibrium points. That is, there will not be a sum over all possible values of some distribution over $i$; there will instead be a single probability distribution $p$ and a single probability distribution $q$, after each has reached free energy minimization.

%
\subsection{Rewriting the Bayesian Posterior Distribution}
\label{subsec:rewriting-Bayesian-posterior}
%

Before we rewrite the reverse K-L divergence term of Eqn.~\ref{eqn:variational-K-L-divergence-summation}, we first recall how the Bayesian posterior probability density can be rewritten, as framed in Blei et al.  \cite{Blei-et-al_2016_Variational-Bayes}.  

Consider a system that has a set of observable variables $\textbf{v}=v_{1..V}$ and a set of latent or ``hidden'' variables $\textbf{w}=w_{1..W}$. In a feedforward neural network, for example, the observable variables $\textbf{v}$ would be the values of the output layer neurons, and the latent (hidden) variables would be the associated values of the hidden layer $\textbf{w}$ neurons. 

Similarly, we can envision many other situations in which we can identify an observation that is a function of multiple input factors. Sometimes, not all of those input factors can be directly observed. 

In the Bayesian formalism, the prior density of the (set of) latent variables $w$ is defined as $p(w)$. A Bayesian model relates these latent variables to the observations $v$ through the likelihood $p(v|w)$. The interpretation is straightforward; it speaks to the likelihood of observing an outcome or observable variable $v$ given the hidden variables $w$. This is called the \textit{prior distribution}. 

Sometimes, though, we don't have an accurate means of establishing the values for the latent or hidden variables $w$. Thus, we use approximate inference to determine the posterior distribution, $p(w|v)$. This means that we are trying to estimate the values of the hidden variables, seeing only the values for the observable variables. 
  
To rewrite the probability density, we first consider a system that can be described in terms of a joint density of latent variables $\textbf{w}=w_{1..W}$ and observations $\textbf{v}=v_{1..V}$, where the conditional density function is given as

\begin{equation}
\label{eqn:conditional-density-func-latent-observables-appendix}
  \begin{aligned}
   p(w|v)=p(w,v)/p(v).    
  \end{aligned}
\end{equation}

Conversely, we also have

\begin{equation}
\label{eqn:joint-density-func-latent-observables-appendix}
  \begin{aligned}
   p(w,v)=p(w|v)p(v).    
  \end{aligned}
\end{equation}

%
\subsection{Rewriting the Reverse K-L Divergence}
\label{subsec:rewriting-KL-divergence}
%

We wish now to rewrite the probability density of the external states $\tilde{\psi}$ that is conditional on the Markov blanket and internal (representational) states, so that the probability density becomes a joint distribution. 

We identify the conditional distribution from Eqn.~\ref{eqn:variational-K-L-divergence-summation} in terms of the joint probability distribution $p(\tilde{{\psi}},\tilde{s},\tilde{a},\tilde{r})$, together with the simple probability distribution over the model states. 

\begin{equation}
\label{eqn:conditional-density-system-terms}
  \begin{aligned}
  p(\tilde{{\psi}}|\tilde{s},\tilde{a},\tilde{r}) =
   p(\tilde{{\psi}},\tilde{s},\tilde{a},\tilde{r})/   
    p(\tilde{s},\tilde{a},\tilde{r}).
  \end{aligned}
\end{equation}

Substituting this result into Eqn.~\ref{eqn:variational-K-L-divergence-summation}, we have

\begin{equation}
\label{eqn:variational-K-L-divergence-summation-rewrite}
  \begin{aligned}
   D_{KL}[q({\tilde{{\psi}}}|\tilde{r})||
    p(\tilde{{\psi}}|\tilde{s},\tilde{a},\tilde{r})] = \sum_{i=1}^I q({\tilde{{\psi}}}|\tilde{r}) \ln\left({\frac {q({\tilde{{\psi}}}|\tilde{r})p(\tilde{s},\tilde{a},\tilde{r})}
    {p(\tilde{{\psi}},\tilde{s},\tilde{a},\tilde{r})}}\right),    
  \end{aligned}
\end{equation}

\noindent which we can reorganize to write as

\begin{equation}
\label{eqn:variational-K-L-divergence-summation-rewrite-2}
  \begin{aligned}
   D_{KL}[q({\tilde{{\psi}}}|\tilde{r})||
    p(\tilde{{\psi}}|\tilde{s},\tilde{a},\tilde{r})]  = \sum_{i=1}^I q({\tilde{{\psi}}}|\tilde{r}) \ln\left( {p(\tilde{s},\tilde{a},\tilde{r})}\right)
\\ + \sum_{i=1}^I q({\tilde{{\psi}}}|\tilde{r}) \ln\left({\frac {q({\tilde{{\psi}}}|\tilde{r})}
    {p(\tilde{{\psi}},\tilde{s},\tilde{a},\tilde{r})}}\right).    
  \end{aligned}
\end{equation}

Following Beal \cite{Beal_2003_Variational-algorithm-approx-Bayes-inference} (Eqns. 2.32 - 2.34), we note that the sum over the model terms $q$ in the first term on the RHS comes to 1 (implicitly, there is a double summation there, and $q$ is independent of $p$), so that we have 

\begin{equation}
\label{eqn:variational-K-L-divergence-summation-rewrite-3}
  \begin{aligned}
   D_{KL}[q({\tilde{{\psi}}}|\tilde{r})||
    p(\tilde{{\psi}}|\tilde{s},\tilde{a},\tilde{r})]  = \sum_{i=1}^I \ln\left( {p(\tilde{s},\tilde{a},\tilde{r})}\right)
\\ + \sum_{i=1}^I q({\tilde{{\psi}}}|\tilde{r}) \ln\left({\frac {q({\tilde{{\psi}}}|\tilde{r})}
    {p(\tilde{{\psi}},\tilde{s},\tilde{a},\tilde{r})}}\right),    
  \end{aligned}
\end{equation}
 
\noindent
keeping in mind that the $q$ are taken over the hidden or latent variables, and that the sum over $i=1..I$ is taken over the $I$ units in the system being modeled.  

This is a good place in which to note that Friston typically writes the first term on the RHS of \ref{eqn:variational-K-L-divergence-summation-rewrite-3} without the summation sign; e.g., the summation is subsumed into the notation. 

For example, Friston, as Eqn. 2.8 in Friston (2013) \cite{Friston_2013_Life-as-we-know-it} and Eqns. 3.2 and 3.4 of Friston et al. (2015) \cite{Friston-et-al_2015_Knowing-ones-place-free-energy-pattern-recognition}, uses

\begin{equation}
\label{eqn:L-thermodynamic-free-energy_third-time}
  \begin{aligned}
    L(\tilde{s},\tilde{a},\tilde{r})
 = -  \ln {p(\tilde{s},\tilde{a},\tilde{r})},   
  \end{aligned}
\end{equation} 

\noindent
where clearly the meaning (see Eqn.~\ref{eqn:variational-K-L-divergence-summation-rewrite-3}) is 

\begin{equation}
\label{eqn:L-thermodynamic-free-energy_third-time-Friston-including-summation}
  \begin{aligned}
    L(\tilde{s},\tilde{a},\tilde{r})
 = -  \sum_{i=1}^I \ln {p(\tilde{s},\tilde{a},\tilde{r})}.   
  \end{aligned}
\end{equation} 

As further evidence that Friston intends the summation (or, as suitable for the situation, an integration) is found in Friston's expression (2013, see Eqn. 2.7) \cite{Friston_2013_Life-as-we-know-it} for $F(\tilde{s},\tilde{a},\tilde{r})$, as 

\begin{equation}
\label{eqn:free-energy-integral}
  \begin{aligned}
    F(\tilde{s},\tilde{a},\tilde{r})
     = - \int_{\psi} q({\tilde{{\psi}}}|\tilde{r}) \ln\left({\frac 
    {p(\tilde{{\psi}},\tilde{s},\tilde{a},\tilde{r}|m)}
     {q({\tilde{{\psi}}}|\tilde{r})}}
    \right)d{\psi}.     
  \end{aligned}
   \nonumber
\end{equation}

We will discuss this equation in greater context later in this Subsection.  

Returning to our original line of thought, we rearrange terms in Eqn.~\ref{eqn:variational-K-L-divergence-summation-rewrite-3} to obtain 

\begin{equation}
\label{eqn:variational-K-L-divergence-summation-rewrite-4}
  \begin{aligned}
    \sum_{i=1}^I q({\tilde{\psi}}|\tilde{r})	 
	\ln\left({\frac {q({\tilde{{\psi}}}|\tilde{r})} 
    {p(\tilde{{\psi}},\tilde{s},\tilde{a},\tilde{r})}}\right) 
    =  - \sum_{i=1}^I \ln\left( {p(\tilde{s},\tilde{a},  \tilde{r})}\right) 
\\    +  D_{KL}[q({\tilde{{\psi}}}|\tilde{r})||           p(\tilde{{\psi}}|\tilde{s},\tilde{a},\tilde{r})].
  \end{aligned}
\end{equation}

Adopting Friston's notation, in which the summation (or integration, in the case of continuous variables) is subsumed, we can write

\begin{equation}
\label{eqn:variational-K-L-divergence-rewrite-4}
  \begin{aligned}
     q({\tilde{\psi}}|\tilde{r})	 
	\ln\left({\frac {q({\tilde{{\psi}}}|\tilde{r})} 
    {p(\tilde{{\psi}},\tilde{s},\tilde{a},\tilde{r})}}\right) 
    =  - \ln\left( {p(\tilde{s},\tilde{a},  \tilde{r})}\right) 
\\    +  D_{KL}[q({\tilde{{\psi}}}|\tilde{r})||           p(\tilde{{\psi}}|\tilde{s},\tilde{a},\tilde{r})].
  \end{aligned}
\end{equation}

Note that the following equations are being written in Friston's style, with summation (or integration) subsumed. 

We notice that Eqn.~\ref{eqn:variational-K-L-divergence-rewrite-4} is similar to form of Eqn.~\ref{eqn:var-free-energy-eqn_part2-first-time-second-half}; sufficiently so that we can establish the identities for $F(\tilde{s},\tilde{a},\tilde{r})$ and $L(\tilde{s},\tilde{a},\tilde{r})$.

For the Left-Hand-Side (LHS) of Eqn.~\ref{eqn:variational-K-L-divergence-rewrite-4}, we create the identity for $F(\tilde{s},\tilde{a},\tilde{r})$ as

\begin{equation}
\label{eqn:free-energy-repeat}
  \begin{aligned}
    F(\tilde{s},\tilde{a},\tilde{r})
 = q({\tilde{{\psi}}}|\tilde{r}) \ln\left({\frac {q({\tilde{{\psi}}}|\tilde{r})}
    {p(\tilde{{\psi}},\tilde{s},\tilde{a},\tilde{r})}}
    \right)    
\\     = - q({\tilde{{\psi}}}|\tilde{r}) \ln\left({\frac 
    {p(\tilde{{\psi}},\tilde{s},\tilde{a},\tilde{r})}
     {q({\tilde{{\psi}}}|\tilde{r})}}
    \right).   
  \end{aligned}
\end{equation}

This gives us the precise form for $F(\tilde{s},\tilde{a},\tilde{r})$; the \textit{variational free energy}. We note the specific difference between this Eqn.~\ref{eqn:free-energy-repeat} and Eqn.~\ref{eqn:variational-K-L-divergence-summation}; both have the form of a reverse K-L divergence. However, in Eqn.~\ref{eqn:free-energy-repeat}, the divergence is between the distribution $q$ of the external system and the joint model-distribution of both the external system and the internal system; in  Eqn.~\ref{eqn:variational-K-L-divergence-summation}, the divergence is between the model distribution $q$ and the (observed) distribution of the external system as conditioned on the representational system and the Markov blanket. 

We notice also (in the second part of Eqn.~\ref{eqn:free-energy-repeat}) that Friston et al. prefer to represent the variational free energy as the \textit{negative} of a divergence-like term; it is now between the joint probability distribution (of the external system observatioin) against the model, although the multiplying factor is still that of the model distribution.  

Similarly, for the first term on the RHS of Eqn.~\ref{eqn:variational-K-L-divergence-rewrite-4}, we also take note of the interpretation offered by Friston (2015) \cite{Friston-et-al_2015_Knowing-ones-place-free-energy-pattern-recognition}, which gives us

\begin{equation}
\label{eqn:L-thermodynamic-free-energy_third-time}
  \begin{aligned}
    L(\tilde{s},\tilde{a},\tilde{r})
 = -  \ln {p(\tilde{s},\tilde{a},\tilde{r})},   
  \end{aligned}
\end{equation} 

\noindent so that $L$ is defined as the \textit{negative} of the logarithm of the probability of internal (representational) units, together with the Markov blanket units. 

As another notational note; Friston actually incorporates dependence on the model parameters into this term; see Eqn. 3.1 of Friston et al. (2015) \cite{Friston-et-al_2015_Knowing-ones-place-free-energy-pattern-recognition} and Eqn. 2.7 of Friston (2013) \cite{Friston_2013_Life-as-we-know-it} for $F(\tilde{s},\tilde{a},\tilde{r})$, which gives 

\begin{equation}
\label{eqn:L-thermodynamic-free-energy_third-time}
  \begin{aligned}
    L(\tilde{s},\tilde{a},\tilde{r})
 = -  \ln {p(\tilde{s},\tilde{a},\tilde{r}|m)},   
  \end{aligned}
  \nonumber
\end{equation}

If we were to substitute these two expressions into Eqn.~\ref{eqn:variational-K-L-divergence-rewrite-4}, we would obtain

\begin{equation}
\label{eqn:variational-free-energy-Gibbs-and-KL}
  \begin{aligned}
    F(\tilde{s},\tilde{a},\tilde{r}) = - q({\tilde{{\psi}}}|\tilde{r}) \ln\left({\frac 
    {p(\tilde{{\psi}},\tilde{s},\tilde{a},\tilde{r})}
     {q({\tilde{{\psi}}}|\tilde{r})}}
    \right)  
\\    =  L(\tilde{s},\tilde{a},\tilde{r}) 
 +  D_{KL}[q({\tilde{{\psi}}}|\tilde{r})||           p(\tilde{{\psi}}|\tilde{s},\tilde{a},\tilde{r})],
  \end{aligned}
\end{equation}

\noindent
which is identical with the second part of  Eqn.~\ref{eqn:var-free-energy-eqn_part2-first-time-second-half}, and also with Friston (2015), Eqn. 3.2 \cite{Friston-et-al_2015_Knowing-ones-place-free-energy-pattern-recognition}, and also with Friston (2013) Eqn. 2.8 \cite{Friston_2013_Life-as-we-know-it}. 

Once again, it may be useful to consider the distinction between $q$ and $L$. More precisely, we need to ask ourselves exactly what it is that we mean when we speak of $p(\tilde{s},\tilde{a}, \tilde{r})$. This is presumably the probability distribution of the internal (representational) and Markov blanket states. However, we have been representing the distribution of the internal and Markov blanket states as $p(x|\theta)$; that is, as a probability distribution of the \textit{observable, representational} system that is encoded by the internal states $\tilde{s},\tilde{a},\tilde{r}$ together with a (set of) model parameters $\theta$. 

A useful interpretation is that we may take $L(\tilde{s},\tilde{a},\tilde{r})$ to be the actual distribution of the representational system (as observed directly over its various components), and $p$ to be the computational distribution of the representation. In short, the negative log likelihood of sensory states (and active plus internal states, i.e.,  $L(\tilde{s},\tilde{a},\tilde{r})$) pertains to the actual state of affairs, while the free energy corresponds to the equivalent measure that would be obtained if the sensory units were caused by the latent or hidden states encoded by the internal states. By minimizing free energy, the two become close but (in general) will never be exactly the same.

The physical implication here could be that we will obtain $p$ as the probability distribution for the observations in a free energy-minimized state. In contrast, the values of specific elements in the distribution over $(\tilde{s},\tilde{a},\tilde{r})$ may, at a certain point, not be in a free energy-minimized state. 

We have thus accomplished half of our goal, in deriving one of the equalities of Eqn.~\ref{eqn:var-free-energy-eqn_part2-first-time}, involving the negative log-likelihood of the probability of states that are actually observed, added to the reverse K-L divergence of the probability distribution of the model with respect to the probability distribution of the external units. 

Examining Eqn.~\ref{eqn:free-energy-repeat}, we note the correspondence between the expression for $F(\tilde{s},\tilde{a},\tilde{r})$ given there and the corollary expression given  for the free energy of a system in Friston (2013, see Eqn. 2.7) \cite{Friston_2013_Life-as-we-know-it}, as

\begin{equation}
\label{eqn:free-energy-integral}
  \begin{aligned}
    F(\tilde{s},\tilde{a},\tilde{r})
     = - \int_{\psi} q({\tilde{{\psi}}}|\tilde{r}) \ln\left({\frac 
    {p(\tilde{{\psi}},\tilde{s},\tilde{a},\tilde{r})}
     {q({\tilde{{\psi}}}|\tilde{r})}}
    \right)d{\psi}.     
  \end{aligned}
\end{equation}

We take note that the integration in Eqn.~\ref{eqn:free-energy-integral} is over the external units $\tilde{\psi}$, reinforcing our understanding that the free energy $F(\tilde{s},\tilde{a},\tilde{r})$ is really the variational free energy of the external system, with the probability distribution in the numerator of the logarithmic term being taken over the joint distribution of external units $\psi$ together with the internal or representational units $\tilde{r}$ as well as the Markov blanket units. 
  
We notice the correspondence between Eqn.~\ref{eqn:variational-free-energy-Gibbs-and-KL} above and Eqn. 2.34 of Beal \cite{Beal_2003_Variational-algorithm-approx-Bayes-inference}, in that $F(\tilde{s},\tilde{a},\tilde{r}) = -F(q_x(x),\theta)$, where the former notation for the free energy is Friston's, and the second is Beal's. (A minor technical note, is that the free energy of a probability distribution ($q$) is a functional, while the free energy of its sufficient statistics ($\tilde{r}$) becomes a function.) Correspondingly, Eqn.~\ref{eqn:variational-free-energy-Gibbs-and-KL} is precisely the negative of Eqn. 2.34 in Beal.  

Now, we wish to show the first part of  Eqn.~\ref{eqn:var-free-energy-eqn_part2-first-time}, which expresses the same free energy in terms of an expectation of log-likelihood of a certain term (whose precise nature will be clarified) minus the entropy of the model as applied to the external units (which will also need to be verified).

%
\section{The Variational Free Energy: Log-Likelihood Expectation and Entropy}
\label{sec:var-FE-log-likelihood-expectation-and-entropy}
%

The previous Subsection~\ref{subsec:rewriting-KL-divergence} presented a derivation for the second half of Eqn.~\ref{eqn:var-free-energy-eqn_part2-first-time}, giving an expression for the variational free energy in terms of the (negative of the) log-likelihood (over the representational system) and the K-L divergence (of the model vis-\`{a}-vis the external system). In this section, we show how the other equality expressed in Eqn.~\ref{eqn:var-free-energy-eqn_part2-first-time} can be derived, giving the free energy in terms of what Friston calls an ``expected enthalpy'' and an entropy term, the exact natures of which will be determined as we proceed \cite{Friston-et-al_2015_Knowing-ones-place-free-energy-pattern-recognition}. 

Several sources remark that the while the first term in the second line of Eqn.~\ref{eqn:var-free-energy-eqn_part2-first-time} is computable, the second term (the reverse K-L divergence) is not  \cite{Beal_2003_Variational-algorithm-approx-Bayes-inference, Blei-et-al_2016_Variational-Bayes}. This then motivates the expression given on the first line of the equation, with the intention of giving an alternative - and computable - formulation for the variational free energy $F$. 

\textbf{\textit{It is worth noting what these various sources say.}} 

\textbf{\textit{Friston et al. (2015) \cite{Friston-et-al_2015_Knowing-ones-place-free-energy-pattern-recognition}}} states (p. 3, just before Lemma 3.1), with regard to the Lagrangian $ L(\tilde{s},\tilde{a},\tilde{r})$ that ``Although we know this Lagrangian exists, it is practically (almost) impossible to evaluate its form. However, there is an alternative formulation of equation (3.1) that allows one to describe the flow in terms of a probabilistic model of how a system thinks it should behave.'' (This is given prior to Friston's Eqn. 3.2, which is our Eqn.~\ref{eqn:var-free-energy-eqn_part2-first-time}, and which motivates the need for that equation.) ... ``The solution to equation (3.2) implies the
internal states minimize free energy rendering the divergence zero (by Gibbs inequality) ... In short, the internal states will appear to engage in Bayesian inference, effectively inferring the (external) causes of sensory states. Furthermore, the active states are complicit in this inference, sampling sensory states that maximize model evidence: in
other words, selecting sensations that the system expects. This is active inference, in which internal states and action minimize free energy—or maximize model evidence—in a way
that is consistent with the good regulator theorem and related treatments of self-organization [49,58–61].''

\textbf{\textit{Beal (2003, p. 45) \cite{Beal_2003_Variational-algorithm-approx-Bayes-inference}}} states that: ``A more principled approach is to estimate the integral numerically by evaluating the integrand at many different $\theta$ via Monte Carlo methods. In the limit of an infinite number of samples of $\theta$ this produces an accurate result, but despite ingenious attempts to curb the curse of dimensionality in $\theta$ using methods such as Markov chain Monte Carlo, these methods remain prohibitively computationally intensive in interesting models. These methods were reviewed in the last chapter, and the bulk of this chapter concentrates on a third way of approximating the integral, using variational methods. The key to the variational method is to approximate the integral with a simpler form that is tractable, forming a lower or upper bound. The integration then translates into the implementationally simpler problem of bound optimisation: making the bound as tight as possible to the true value.''

\textbf{\textit{Blei et al. (2018, p. 2) \cite{Blei-et-al_2016_Variational-Bayes}}}  state that: ``For decades, the dominant paradigm for approximate inference has been MCMC [Markov chain Monte Carlo] ... However, there are problems for which we cannot easily use this approach. These arise particularly when we need an approximate conditional faster than a simple MCMC algorithm can produce, such as when data sets are large or models are very complex. In these settings, variational inference provides a good alternative approach to approximate Bayesian inference. Rather than use sampling, the main idea behind variational inference is to use optimization. First, we posit a family of approximate densities $Q$. This is a set of densities over the latent variables. Then, we try to find the member of that family that minimizes the Kullback-Leibler (KL) divergence to the exact posterior ...  Finally, we approximate the posterior with the optimized member of the family $q^*(\cdot)$).

\textbf{\textit{Of these various explanations, that of Blei et al. seems most straight-forward. }} 

\textbf{\textit{We wish to minimize the [reverse] Kullback-Leibler divergence of Eqn.~\ref{eqn:var-free-energy-eqn_part2-first-time}.}}

The derivation of the first part of Eqn.~\ref{eqn:var-free-energy-eqn_part2-first-time} is found in Beal (2003) \cite{Beal_2003_Variational-algorithm-approx-Bayes-inference}, Eqns. 2.15 and 2.16. 

For convenience, Eqn.~\ref{eqn:var-free-energy-eqn_part2-first-time} (Eqn. 3.2 in Friston et al. (2015) \cite{Friston-et-al_2015_Knowing-ones-place-free-energy-pattern-recognition}) is presented again, as

\begin{equation}
\label{eqn:var-free-energy-eqn_part2-restated}
  \begin{aligned}
  F(\tilde{s},\tilde{a},\tilde{r}) =
    E_q[L(\tilde{x})]
    - H[q(\tilde{{\psi}}|\tilde{r})] \\
    = L(\tilde{s},\tilde{a},\tilde{r}) + 
    D_{KL}[q({\tilde{{\psi}}}|\tilde{r})||
    p(\tilde{{\psi}}|\tilde{s},\tilde{a},\tilde{r})]. \nonumber    
  \end{aligned}
\end{equation}

Our goal is to verify the first equality presented in this equation. To accomplish this, we follow a line of reasoning presented in Beal (2003) \cite{Beal_2003_Variational-algorithm-approx-Bayes-inference}, who introduced a formulation for the log likelihood.

We begin with Beal's Eqn. 2.10, given as 

\begin{equation}
\label{eqn:Beal-eqn-2-10}
  \begin{aligned}
    L(\theta) & \equiv \ln p(y|\theta) 
    = \sum_{i=1}^n \ln p(y_i|\theta)
    = \sum_{i=1}^n \ln \int dx_i \: p(x_i,y_i|\theta) .    
  \end{aligned}
\end{equation}

Beal's Eqns. 2.12 - 2.16 are reproduced here as

\begin{equation}
\label{eqn:basic-free-energy-Beal-restated-main-text}
  \begin{aligned}
    L(\theta) 
&    = \sum_{i=1} \ln \int dx_i \: p(x_i,y_i|\theta) \\
&    =  \sum_{i=1} \ln \int dx_i \: q_{x_i}(x_i) \: \frac { p(x_i,y_i|\theta)} {q_{x_i}{(x_i})}  \\
&    \geq  \sum_{i=1} \int dx_i \: q_{x_i}(x_i) \: \ln \: \frac { p(x_i,y_i|\theta)} {q_{x_i}{(x_i})} \\
&    =  \sum_{i=1} \left( \int dx_i \: q_{x_i}(x_i) \: \ln  \: p(x_i,y_i|\theta) - \int dx_i \:  q_{x_i} (x_i) \: \ln \: q_{x_i}(x_i) \right) \\
&    =  \sum_{i=1} \left( \int dx_i \: q_{x_i}(x_i) \: \ln  \: p(x_i,y_i|\theta) \right)  - \int dx_i \:  q_{x_i} (x_i) \: \ln \: q_{x_i}(x_i) \\
& \equiv F(q_{x_1}{(x_1}),..., q_{x_n}{(x_n}), \theta).    
  \end{aligned}
\end{equation}

In the preceding Eqn.~\ref{eqn:basic-free-energy-Beal-restated-main-text}, we note that the last two lines are those that interest us; we see there a formal similarity between those terms and those in the first equality expression of Eqn.~\ref{eqn:var-free-energy-eqn_part2-first-time}. Specifically, we desire to show a correspondence between Beal's Eqns. 2.12 - 2.16, given as 

\begin{equation}
\label{eqn:basic-free-energy-Beal-restated-main-text-condensed}
  \begin{aligned}
    L(\theta) 
&    =  \sum_{i=1} \ln \int dx_i \: q_{x_i}(x_i) \: \frac { p(x_i,y_i|\theta)} {q_{x_i}{(x_i})}  \\
&    \geq  \sum_{i=1} \left( \int dx_i \: q_{x_i}(x_i) \: \ln  \: p(x_i,y_i|\theta) \right)  - \int dx_i \:  q_{x_i} (x_i) \: \ln \: q_{x_i}(x_i) \\
& \equiv F(q_{x_1}{(x_1}),..., q_{x_n}{(x_n}), \theta),    
  \end{aligned}
\end{equation}

\noindent
and Friston's equation, which we've presented here as Eqn.~\ref{eqn:var-free-energy-eqn_part2-first-time} (Eqn. 3.2 in Friston et al. (2015), and which we present again as

\begin{equation}
\label{eqn:var-free-energy-eqn_part2-first-line-only-again}
  \begin{aligned}
  F(\tilde{s},\tilde{a},\tilde{r}) =
    E_q[L(\tilde{x})]
    - H[q(\tilde{{\psi}}|\tilde{r})].    
  \end{aligned}
\end{equation}

The ``greater-than-or-equal'' relation in Eqn.~\ref{eqn:basic-free-energy-Beal-restated-main-text-condensed} is due to Jensen's inequality, and is essential to one of the steps shown in Eqn.~\ref{eqn:basic-free-energy-Beal-restated-main-text}; this is a minor omission in Friston's phrasing, and does not substantially impact our translation. 

We will note, as we pursue our investigation, that Friston's identification of $F$ is the negative of that used by Beal; this will change the direction of the inequality, but will again not impact our work. 

As a precursor step, we use the equality case and take negative of Beal's expression from Eqn.~\ref{eqn:basic-free-energy-Beal-restated-main-text-condensed} and write

\begin{equation}
\label{eqn:basic-free-energy-Beal-restated-main-text-condensed-negative-both-sides}
  \begin{aligned}
    F(\tilde{s},\tilde{a},\tilde{r}) \\
& =    - F(q_{x_1}{(x_1}),..., q_{x_n}{(x_n}), \theta) \\
& =  - \sum_{i=1} \left( \int dx_i \: q_{x_i}(x_i) \: \ln  \: p(x_i,y_i|\theta) \right)  \\ 
& - \left[ - \int dx_i \:  q_{x_i} (x_i) \: \ln \: q_{x_i}(x_i) \right]. 
  \end{aligned}  
\end{equation}

We identify the three separate equivalences that we will wish to make, correlating Friston's terms (Eqn.~\ref{eqn:desired-equiv-for-F-again}) with Beal's (Eqn.~\ref{eqn:basic-free-energy-Beal-restated-main-text-condensed-negative-both-sides}); specifically that

\begin{equation}
\label{eqn:desired-equiv-for-F-again}
  \begin{aligned}
  F(\tilde{s},\tilde{a},\tilde{r}) \\
& =  - F(q_{x_1}{(x_1}),..., q_{x_n}{(x_n}), \theta) \\  
&  = - \sum_{i=1} \ln \int dx_i \: q_{x_i}(x_i) \: \frac { p(x_i,y_i|\theta)} {q_{x_i}{(x_i})},    
  \end{aligned}
\end{equation}

and

\begin{equation}
\label{eqn:desired-equiv-for-L}
  \begin{aligned}
    E_q[L(\tilde{x})] = - \sum_{i=1} \left( \int dx_i \: q_{x_i}(x_i) \: \ln  \: p(x_i,y_i|\theta) \right),     
  \end{aligned}
\end{equation}

and

\begin{equation}
\label{eqn:desired-equiv-for-H}
  \begin{aligned}
   H[q(\tilde{{\psi}}|\tilde{r})] =  
     - \int dx_i \:  q_{x_i} (x_i) \: \ln \: q_{x_i}(x_i) .    
  \end{aligned}
\end{equation}

We remind ourselves (to avoid confusion for any who would be reading and comparing the original documents) that Friston's $F(\tilde{s},\tilde{a},\tilde{r})$ is the \textit{negative} of Beal's $ F(q_{x_1}{(x_1}),..., q_{x_n}{(x_n}), \theta)$.

We will address each of these three correspondences in the following three subsections, respectively.

%
\subsection{Equivalence of the Variational Free Energy Expressions}
\label{subsec:equiv-var-free-energy}
%

In this subsection, we will show the correspondence given previously as Eqn.~\ref{eqn:desired-equiv-for-F-again}, that is

\begin{equation}
\label{eqn:desired-equiv-for-F-again-second-time}
  \begin{aligned}
  F(\tilde{s},\tilde{a},\tilde{r}) \\
& =  - F(q_{x_1}{(x_1}),..., q_{x_n}{(x_n}), \theta) \\  
&  = - \sum_{i=1} \ln \int dx_i \: q_{x_i}(x_i) \: \frac { p(x_i,y_i|\theta)} {q_{x_i}{(x_i})}.    
  \end{aligned}
\nonumber  
\end{equation}

We pause to recollect, from Subsection~\ref{subsec:rewriting-KL-divergence}, Eqn.~\ref{eqn:free-energy-repeat}, the term that Friston has identified for variational free energy. We state this again here for reference as

\begin{equation}
\label{eqn:free-energy-summation-repeat-2}
  \begin{aligned}
    F(\tilde{s},\tilde{a},\tilde{r})
 = q({\tilde{{\psi}}}|\tilde{r}) \ln\left({\frac {q({\tilde{{\psi}}}|\tilde{r})}
    {p(\tilde{{\psi}},\tilde{s},\tilde{a},\tilde{r})}}
    \right)    
\\     = - q({\tilde{{\psi}}}|\tilde{r}) \ln\left({\frac 
    {p(\tilde{{\psi}},\tilde{s},\tilde{a},\tilde{r})}
     {q({\tilde{{\psi}}}|\tilde{r})}}
    \right).   
  \end{aligned}
\end{equation}

We rewrite Beal's expresssion from Eqn.~\ref{eqn:basic-free-energy-Beal-restated-main-text-condensed}, by taking the negative of all terms and dropping the inequality 

\begin{equation}
\label{eqn:basic-free-energy-Beal-restated-main-text-condensed-more}
  \begin{aligned}
    -L(\theta) &  = - \sum_{i=1} \ln \int dx_i \: q_{x_i}(x_i) \: \frac { p(x_i,y_i|\theta)} {q_{x_i}{(x_i})} \\
    &= -F(q_{x_1}{(x_1}),..., q_{x_n}{(x_n}), \theta) .    
  \end{aligned}
\end{equation}

We note that there is indeed the desired resemblance. To be more clear, we are seeking the correspondence that can be expressed as

\begin{equation}
\label{eqn:correspondence-Beal-Friston-var-energy-term}
  \begin{aligned}
    q({\tilde{{\psi}}}|\tilde{r}) \ln\left({\frac 
    {p(\tilde{{\psi}},\tilde{s},\tilde{a},\tilde{r})}
     {q({\tilde{{\psi}}}|\tilde{r})}}
    \right) \\   
    &  = \sum_{i=1} \ln \int dx_i \: q_{x_i}(x_i) \: \frac { p(x_i,y_i|\theta)} {q_{x_i}{(x_i})}.   
  \end{aligned}
\end{equation}

Clearly, the desired terms are present and in their appropriate order. The key differences are that Beal (on the Right-Hand-Side, or RHS) explicitly identifies the summation and integration steps, and that the Friston formalism is expanded; it includes the entire set of elements; $\tilde{\psi}$, $\tilde{s}$, $\tilde{r}$, and $\tilde{r}$. 

This makes sense; the entire ``universe'' encompassed in the Friston model is expressed via  $\tilde{\psi}$, $\tilde{s}$, $\tilde{r}$, and $\tilde{r}$. Correspondingly, the ``universe'' modeled in Beal's approach is the set of observable data points $y_i$ and the associated latent variables $x_i$. The parameter $\theta$ is expressly identified in Beal's notation; it is suppressed in this particular notation by Friston, but is evident in various Friston writings (op. cit.). 

We give our attention to how Beal's expression integrates over the $q(x_i)$. More precisely, Beal gives an integration over the hidden or latent elements $x_i$, and a summation over the units $y_i$. Friston's approach simply specifies a distribution $q$ associated with each specific probabilistic state $p$. However, as discussed earlier, Friston's notation subsumes the summation (or integration, as appropriate). 

This is a reasonable transition, as in the system being described by Friston, we no longer are assessing the values of $p$ and $q$ over the same underlying hidden variables $x$. Rather, the $q$ corresponds to the external system, and the $p$ corresponds to the internal (representational) system, which we are bringing into alignment with the external system.

%
\subsection{Equivalence of the Entropy Expressions}
\label{subsec:equiv-entropy}
%

For ease in flow, we next address the equivalence of the two entropy terms, as this is relatively straightforward. 

Friston identifies an entropy term $H$ (using this notation, common to information theory, rather than the more classic thermodynamic notation $S$), and we desire that it be equivalent to Beal's term, as expressed previously in Eqn.~\ref{eqn:desired-equiv-for-H}, which we restate here (using Beal's notation) as 

\begin{equation}
\label{eqn:desired-equiv-for-H-repeat}
  \begin{aligned}
   H[q(\tilde{{\psi}}|\tilde{r})] =  
     - \int dx \:  q_{x} (x) \: \ln \: q_{x}(x), \nonumber    
  \end{aligned}
\end{equation}

\noindent
which we can also write (using Friston's notation) as

\begin{equation}
\label{eqn:desired-equiv-for-H-repeat}
  \begin{aligned}
   H[q(\tilde{{\psi}}|\tilde{r})] =  
     - \int d({\tilde{{\psi}}}|\tilde{r}) \: q({\tilde{{\psi}}}|\tilde{r}) \: \ln \: q({\tilde{{\psi}}}|\tilde{r}) . \nonumber    
  \end{aligned}
\end{equation}

We recall that the fundamental definition for the entropy of a system is given (see Appendix \ref{sec:Appendix-var-FE-enthalpy-entropy}) as

\begin{equation}
\label{eqn:Feynmann-entropy-repeat-fm-Appendix}
  \begin{aligned}
    S =  - k   \sum_{n}  \:  P_n \ln \: P_n,     
  \end{aligned}
\end{equation}

\noindent
where $P_n$ refers to the probability of a unit being in energy state $n$. This is a classic entropy formulation, and we see it replicated in Eqn.~\ref{eqn:desired-equiv-for-H}. The thing that we wish to carefully note is that in Eqn.~\ref{eqn:desired-equiv-for-H}, given that the entropy is being expressed as a function of $ q({x_i})$ (using Beal's notation) or $q(\tilde{\psi}|\tilde{r})$ (using Friston's notation), the units that are being summed (or integrated) are those in the model-distribution of the external system $\tilde{\psi}$, as conditioned on the units in the internal system $\tilde{r}$. Thus,  $H$ is a function of (the model of) the external system represented by $q$. 

Very specifically, when it comes to evaluating this function, we would not need a distribution over all possible states in the model. Rather, the computational engine which this Technical Report envisions is one in which the external and internal systems separately come to free energy minima. Thus, when that minimum point is achieved, there would be a single existent value for $p$; the one which represents the free energy-minimized state for a given set of parameters $\theta$. This would then lead (via sensory and active units) to a similarly free energy-mininized $q$.

Friston (personal communication) notes that ``A complementary perspective on this computational saving follows from Feynman's original motivation; namely, that we have converted a very difficult integration problem into an easy optimization problem. Here, the optimization problem simply entails minimizing variational free energy.''

%
\subsection{Equivalence of the Enthalpy Expressions}
\label{subsec:equiv-enthalpy}
%

Finally, we wish to show the equivalence between the enthalpy terms. The word ``enthalpy'' may be a misnomer here, but is being used in the classic sense of thermodynamics, in which (see Appendix \ref{sec:Fund-thermo-concepts}) the free energy $F$ is equal to the enthalpy $H$ (which is the classic notation for enthalpy, although $U$ is sometimes used, depending on the version of free energy being described) minus the temperature $T$ times the entropy $S$, or (in the case of this manuscript as well as in Friston's work and most information theoretic works) $H$. 

Thus, the most classic equation in thermodynamics is

\begin{equation}
\label{eqn:basic-eqn-free-energy-classic-repeat-from-Appendix}
  \begin{aligned}
    F =  H \: - \: T S,  \nonumber    
  \end{aligned}
\end{equation} 

\noindent
which states that the free energy is the enthalpy minus temperature times entropy, and where enthalpy is denoted $H$ and entropy is denoted $S$. 

The equations presented by Beal and Friston have the same formal structure as the classic free energy equation from statistical thermodynamics, as stated in Eqn.~\ref{eqn:var-free-energy-eqn_part2-first-time}. The version offered by Friston is replicated here as

\begin{equation}
\label{eqn:var-free-energy-eqn_part2-first-line-only-repeat}
  \begin{aligned}
  F(\tilde{s},\tilde{a},\tilde{r}) =
    E_q[L(\tilde{x})]
    - H[q(\tilde{{\psi}}|\tilde{r})],    
  \end{aligned} \nonumber
\end{equation}

\noindent
and note again that Friston uses $H$ for entropy, instead of the thermodynamic $S$.

The temperature $T$ has been absorbed in the derivations presented in this work; we are dealing with something called a reduced free energy (and also reduced entropy and reduced enthalpy), which are dimensionless quantities. (Note also that this ``reduction'' also normalizes the thermodynamic variables with regard to the total number of units in the system. See Appendix \ref{sec:Fund-thermo-concepts} for a review of basic thermodynamics.)  

We have already established the correspondence between the Friston's variational free energy $F(\tilde{s},\tilde{a},\tilde{r})$ and (the \textit{negative} of) that used by Beal, and also identified it as corresponding (in position and form) to the classical free energy. We have also established the correspondence of Friston's entropy term $H$ with (the \textit{negative} of) that used by Beal, and identified it as corresponding to the classical entropy term. (In this case the actual expressions are very much aligned.) 

We now seek to identify the correspondence between Friston's enthalpy-like term, $E_q[L(\tilde{x})]$, and the \textit{negative} of that used by Beal.  Specifically, we want to establish Eqn.~\ref{eqn:desired-equiv-for-L}, restating it here for convenience as

\begin{equation}
\label{eqn:desired-equiv-for-L-repeat}
  \begin{aligned}
    E_q[L(\tilde{x})] = - \sum_{i=1} \left( \int dx_i \: q_{x_i}(x_i) \: \ln  \: p(x_i,y_i|\theta) \right).      
  \end{aligned}
\end{equation}

We also take note of the interpretation offered by Friston (2015) \cite{Friston-et-al_2015_Knowing-ones-place-free-energy-pattern-recognition}, which states that $L(\tilde{x}) =  -\ln \: {p(\tilde{\psi},\tilde{s},\tilde{a},\tilde{r}|m)}$ (see Lemma 3.1 and also Eqn. 3.2), or more specifically

\begin{equation}
\label{eqn:L-thermodynamic-free-energy-summation_joint-probability}
  \begin{aligned}
    L(\tilde{x}) = L(\tilde{{\psi}},\tilde{s},\tilde{a},\tilde{r})
 = -  \ln {p(\tilde{{\psi}},\tilde{s},\tilde{a},\tilde{r})},   
  \end{aligned}
\end{equation} 

\noindent so that $L$ is defined as the \textit{negative} of the sum of the logarithm of the joint probability of the external and internal (representational) units, together with the Markov blanket units. (The dependence on the model parameter $m$ is implicit.)

We have previously addressed the nature of $L$ in Subsection~\ref{subsec:rewriting-KL-divergence}, and thus will just briefly recapitulate cogent arguments here. 

First, we examine the term $\ln  \: p(x_i,y_i|\theta)$. The joint probability of the dependent variables $y_i$, co-occurring with the independent variables $x_i$, as conditioned by the model system parameters $\theta$, is consistent with Friston's notation involving a joint probability distribution. 

Second, we consider the integration over the $q_{x_i}(x_i)$ times the logarithm of the probability. As noted previously, the $q$ and the $p$ address the distributions over different systems, and thus are independent (to a first order). Thus, we can separate out the integration of the $q$. The agreement with  Eqn.~\ref{eqn:L-thermodynamic-free-energy-summation_joint-probability} becomes self-evident, if we associate the external states with $x$, and the sensory states (plus active and internal states) with $y$.

%
\subsection{Recapitulation and Summary}
\label{subsec:recap-and-summary}
%

We now recast Eqn.~\ref{eqn:basic-free-energy-Beal-restated-main-text} using Friston's notation. 

Further, since  $L(\tilde{{\psi}},\tilde{s},\tilde{a},\tilde{r}) 
= -ln ( {p(\tilde{{\psi}},\tilde{s},\tilde{a},\tilde{r}|m)})$, the signs on the terms on the RHS of  Eqn.~\ref{eqn:basic-free-energy-Beal-restated-main-text} have been changed throughout, along with the direction of the inequality. 

A key feature in the following Eqn.~\ref{eqn:basic-free-energy-Beal-using-Friston-notation} is that Friston is taking the integration over the units $\tilde{\psi}$ in the external system, similar to how Beal is summing over the observable units $y$. However, we will conduct our integration (or summation, which is more literally the case) over the units in the model system, as discussed in Subsection~\ref{subsec:intgrating-over-model}.

Specifically, Friston's starting point is Eqn. 2.7 in \cite{Friston_2013_Life-as-we-know-it}, given as

\begin{equation}
\label{eqn:basic-free-energy-Friston-2015-notation}
  \begin{aligned}
    F(s,a,r)  
&    =  - \int_{\psi} d{\psi} \:q(\psi|r) \ln  \left( \frac {p(\psi,s,a,r)} {q(\psi|r)} \right) \\ 
&    =  E_q[L(\psi, s,a,r)] - H[q(\psi|\mu)]. \nonumber 
  \end{aligned} 
\end{equation} 

Note that the tilde notation, indicating generalized variables, is dropped, conforming with the notation that Friston uses in \cite{Friston_2013_Life-as-we-know-it}, where Friston uses $G$ for $L$ in \cite{Friston_2013_Life-as-we-know-it}. 

Friston's interpretation is that ``Here, free energy is a functional of an arbitrary (variational) density $q(\psi|r)$ [$q(\psi|\lambda)$ in the original article] that is parametrized by internal states. The last equality just shows that free energy can be expressed as the expected Gibbs energy minus the entropy of the variational density.'' (Friston (2013) \cite{Friston_2013_Life-as-we-know-it}, immediately after Eqn. 2.7.)

The corresponding expressions, from Eqns. 2.12 - 2.16 in Beal \cite{Beal_2003_Variational-algorithm-approx-Bayes-inference} are given as

\begin{equation}
\label{eqn:basic-free-energy-Beal-restated-main-text-again}
  \begin{aligned}
    L(\theta) 
&    \geq  \sum_{i=1} \int dx_i \: q_{x_i}(x_i) \: \ln \: \frac { p(x_i,y_i|\theta)} {q_{x_i}{(x_i})} \\
&    =  \sum_{i=1} \left( \int dx_i \: q_{x_i}(x_i) \: \ln  \: p(x_i,y_i|\theta) \right)  - \int dx_i \:  q_{x_i} (x_i) \: \ln \: q_{x_i}(x_i) \\
& \equiv F(q_{x_1}{(x_1}),..., q_{x_n}{(x_n}), \theta). \nonumber   
  \end{aligned}
\end{equation}

The full derivation, using Friston's notation, can be found as

\begin{equation}
\label{eqn:basic-free-energy-Beal-using-Friston-notation}
  \begin{aligned}
    L(s,a,r) 
&= - \ln ( {p(\psi,s,a,r|m)}) \\
&=  - \int_{\psi} d{\psi} \: \ln\left( {p(\psi,s,a,r)}\right) \\   
&=  - \int_{\psi} d{\psi} \: \ln \left( q(\psi|r)\frac {p(\psi, s,a,r)} {q(\psi|r)} \right) \\ 
&    \leq  - \int_{\psi} d{\psi} \:q(\psi|r) \ln  \left( \frac {p(\psi, s,a,r)} {q(\psi|r)} \right) \\ 
&    =  - \int_{\psi} d{\psi} \:q(\psi|r) \ln  \left(  {p(\psi, s,a,r)}  \right) + \int_{\psi} d{\psi} \:q(\psi|r) \ln  \left(  {q(\psi|r)} \right) \\ 
&    =  E_q[G(\psi, s,a,r)] - H[q(\psi|\mu)]\\
& \equiv  F(s,a,r).  
  \end{aligned}
\end{equation} 

A notational point is that in this equation, $G$ refers to the thermodynamic Gibbs free energy, which is being used here in a didactic manner.

As another small note, Friston shifts notation between the third-to-last and the second-to-last lines of this equation, where he expresses the results in Friston (2013) \cite{Friston_2013_Life-as-we-know-it}. In the third-to-last equation, he has the expression involving $q({\tilde{{\psi}}}|\tilde{r})$. In the second-to-last equation, he uses $q({\tilde{{\psi}}}|\mu)$. A rationale is that after the integration (where the units that are being considered in the distributionl $q$ are dependent on the actual representational units $\tilde{r}$), the dependence of $q$ on $\tilde{r}$ no longer needs to be explicitly stated. The introduction of $\mu$ is simply noting that the computation for the distribution $q$ was done with reference to sufficient statistics or parameters $\mu$, which are associated with the internal states ($\mu =\tilde{r}$).

%
\section{Discussion}
\label{sec:discussion}
%

Now that we've done a detailed derivation for both of the equalities expressed in Eqn.~\ref{eqn:var-free-energy-eqn_part2-first-time}, it is useful to step back and ascertain exactly what is meant by these paired statements, which are reproduced below for convenience.

\begin{equation}
\label{eqn:var-free-energy-eqn_part2-reproduced-in-discussion}
  \begin{aligned}
  F(\tilde{s},\tilde{a},\tilde{r}) =
    E_q[L(\tilde{x})]
    - H[q(\tilde{{\psi}}|\tilde{r})] \\
    = L(\tilde{s},\tilde{a},\tilde{r}) + 
    D_{KL}[q({\tilde{{\psi}}}|\tilde{r})||
    p(\tilde{{\psi}}|\tilde{s},\tilde{a},\tilde{r})].  \nonumber  
  \end{aligned}
\end{equation}

The first expression for the variational free energy puts the influence of the external units in the first free energy term $E_q[L(\tilde{x})]$. By stating that we desire the \textit{expectation} of $L(\tilde{x})$, we are pushing to identify $L$ at the point at which we ``expect'' the system to come to a stable state, i.e., a free energy minimum.  

The influence of the ``variation'' or the perturbation to the system is expressed in terms of the ``entropy of the variational density,'' $H[q(\tilde{{\psi}}|\tilde{r})]$. This puts the variation or perturbation of the external units in the context of the expected values for the internal and Markov blanket units. 

The second expression for the variational free energy simply identifies a free energy-like term that involves only internal and Markov blanket units; $L(\tilde{s},\tilde{a},\tilde{r})$. (In statistical terms, this is known as the marginal likelihood; having integrated out dependencies on the external states or causes of sensory states. In Bayesian statistics, this is also known as the (negative) log model evidence (see below).) 

The extracted influence of the expected external units (those typically associated with a specific state of internal and Markov blanket units) is now combined with the influence of the variational (or perturbed) external units, within the reverse Kullback-Leibler divergence term, $D_{KL}[q({\tilde{{\psi}}}|\tilde{r})||     p(\tilde{{\psi}}|\tilde{s},\tilde{a},\tilde{r})]$. 

(In terms of Bayesian statistics, this is the divergence between the approximate and true posterior. This means that minimizing free energy is equivalent to approximate Bayesian inference.) 

This Technical Report has, thus far, served to present in detail a derivation for the ideas behind the variational Bayes approach, and provided a detailed correlation between the notation used by one author (Beal \cite{Beal_2003_Variational-algorithm-approx-Bayes-inference}) and that used by Friston (op. cit.), in his extension of variational Bayes to a more general case, in which an external system is separated from a ``representational'' system by a Markov blanket of sensory and active units. These are two different ways of envisioning the variational Bayes approach in action.

%
\subsection{Free Energy Physical Interpretation}
\label{subsec:free-energy-physical-interpretation}
%

We consider that the free energy formulation that we have been developing describes a system with external units $\psi$, together with a representational system that contains internal units that encode ``latent'' or ``hidden'' states, in terms of their sufficient statistics $\tilde{r}$, that are separated from the external system by a Markov blanket comprising sensory units $\tilde{s}$ and action units $\tilde{a}$, as illustrated in Figure 1 of \cite{Friston_2013_Life-as-we-know-it}. In other words, internal states encode probability distributions over latent states that `could have' caused the sensory states.

Eqn.~\ref{eqn:var-free-energy-eqn_part2-reproduced-in-discussion} gives us the free energy of the system, where the elements of $F(\tilde{s},\tilde{a},\tilde{r})$ are formulated in terms of the probability distribution over $\tilde{\psi}$ in terms of $q({\tilde{\psi}}|\tilde{r})$. In contrast, $L(\tilde{s},\tilde{a},\tilde{r})$ is a function strictly of the units associated with the representation, where the elements include the representational units $\tilde{r}$ along with the Markov blanket units $\tilde{s}$ and $\tilde{a}$. Finally, the reverse K-L divergence term (the final term on the RHS of Eqn.~\ref{eqn:var-free-energy-eqn_part2-reproduced-in-discussion}) expresses the divergence between the model (expressed as $q$) and the representation of the external system (expressed as the posterior distribution $p$), given the Markov blanket. (We are dropping the ``tilde'' notation favored by Friston et al. (op. cit.).)


%
\subsection{Free Energy as a Lower Bound}
\label{subsec:free-energy-lower-bound}
%

Beal notes that $F(q_x(x),\theta)$ is a lower bound on $L(\theta)$ and is a functional of the free
distributions $q_{x_i}(x_i)$ and of $\theta $ (the dependence on $y$ is left implicit). The inequality introduced in the third expression makes use of \textit{Jensen's inequality}. 

Beal notes: ``Defining the energy of a global configuration $(x, y)$ ... the lower bound $F(q_x(x), \theta)\leq L(\theta)$ is the negative of a quantity known in statistical physics as the free energy: the expected energy under $q_x(x)$ minus the entropy of $q_x(x)$ (Feynman, 1972; Neal and Hinton, 1998).'' 

Beal further notes that \textbf{$F(q_x(x),\theta)$ is the negative} of what is known, in statistical thermodynamics, as the \textit{free energy} of a system, which is the \textit{expected energy} ($H$) under $q_{x}(x)$ minus the entropy of $q_{x}(x)$. Thus, when we shift to the notation of Friston (op.cit.), we reverse the signs on all of the terms on the right-hand-side of Eqn.~\ref{eqn:basic-free-energy-Beal-using-Friston-notation}, as well as the direction of the inequality.

As is often noted \cite{Friston_2013_Life-as-we-know-it, Beal_2003_Variational-algorithm-approx-Bayes-inference, Blei-et-al_2016_Variational-Bayes}, since the $D_{KL}>=0$, the free energy for the model is a lower bound for the free energy of the external system. As the model is brought closer to alignment with the external system (the reverse K-L divergence decreases), the free energy of the model approaches that of the external system ($L(\tilde{s},\tilde{a},\tilde{r}) => F(\tilde{s},\tilde{a},\tilde{r})$).

%
\section{The Evolution of Active Inference}
\label{sec:evolution-active-inference}
%

This paper has addressed the notational correspondence between Friston's early work introducing active inference (op. cit.) and the notation that he followed from Beal (2003) \cite{ Beal_2003_Variational-algorithm-approx-Bayes-inference}, with some attention also to notation used by Blei et al. (2016) \cite{Blei-et-al_2016_Variational-Bayes}.

This early Friston work (approximately between 2010 and 2015) emphasized the distinction between the external environment $\Psi$ and the representation of that environment $\tilde{r}$, as mediated by sensing agents $\tilde{s}$ and action agents $\tilde{a}$.

More recently, Friston et al. have shifted notation (starting 2016 - 2017) \cite{Friston-et-al_2016_Active-inference-and-learning, Friston-et-al_2017_Active-inference-process-theory}. 

The newer formulations have been the basis for recent work on active inference, specifically the evolution of Action Perception Divergence (APD) by Hafner et al. (2020, rev. 2022) \cite{Hafner-et-al_2022_Action-perception-divergence}. The newer notation has also been used by Friston et al. in more broadly describing the ``free energy principle'' \cite{Friston-et-al_2023_Free-energy-principle}.

Authors presenting active inference in more readily-understood forms (compared to Friston's early works), as well as Friston himself, emphasize the role of active inference in ``process theory'' - that is, as a guiding framework for how intelligent systems actually accomplish desired tasks. A recent book by Parr, Pezzulo, and Friston (2022) \cite{Parr-et-al_2022_Active-inference-book} is a premier example of such, as is an excellent review by Sajid et al. (2020), presenting a contrast-and-compare between active inference and reinforcement learning \cite{Sajid-et-al_2020_Active-inference-demystified}.

Most recently, Friston et al. (2024), have put forth a ``renormalization group'' approach to active inference, which allows active inference to be applied to larger-scale problems \cite{Friston-et-al_2024_From-pixels-to-planning}. This overcomes a prior drawback to using active inference --  its restriction to relatively small-scale problems -- such as was done in Cullen et al. (2016) \cite{Cullen-et-al_2018_Active-inference}, where active inference was applied to the game of \textit{Doom}.

%
\section{The Variational Free Energy in a New Computational Engine}
\label{sec:var-FE-new-comput-engine}
%

One of the themes that consistently underlies active inference is that a given system will seek to reach a \textit{free energy equilibrium}. Ideally, the external reality that we are seeking to represent, $\Psi$, undergoes its own processes that continually move it towards a free energy-minimized state, while also adapting to inputs within its own environment as well as actions from the internal representation system; both of these can affect exactly \textit{where} the corresponding free energy minimum might be found. 

In a like vein, the internal representation of this external system, $p(s, a, r)$ (tilde notation removed for simplicity) should likewise come to a free energy minimized state. Again, exactly \textit{where} this free energy minimum is located can change, subject to sensory inputs from the external environment, mediated by sensing agents $(s)$.   

CORTECONs(R) (COntent-Retentive, TEmporally-CONnected neural networks) provide a means for actually bringing an internal representation system to a free energy-minimized state. They do this by establishing a grid of bistate nodes; that is, each node can be in state \textbf{A} or state \textbf{B}; ``on'' or ``off.'' This grid of bistate nodes can be brought to an equilibrium by minimizing a free energy equation that is more complex than usual. The interesting and distinctive characteristic of this equation is that the entropy term encompasses not only whether a given node is ``on'' or ``off,'' but also takes note of the distribution of local patterns - nearest-neghbors, next-nearest-neighbors, and triplets. 

In their simplest form, we consider a CORTECON(R) \textit{only} from the perspective of using it to create a 1-D or 2-D grid of nodes that can form a representation of some external system. In this simplest possible CORTECON(R) interpretation, we pay attention only to the grid free energy, and can adjust node activations (flipping nodes between ``on'' and ``off'' states) to achieve a free energy minimum. 

This use, of course, is simply the most basic sort. More comprehensive CORTECON(R) implementations will allow grid nodes to play the role of latent variables, and the degree to which nodes can be made active can be a function of not only direct stimulus from an external source but also of control parameters $(\varepsilon_0, \varepsilon_1)$. 

In the limited treatment that we provide in this paper, we address only the free energy equation that we used in a CORTECON(R), which is taken directly from the cluster variation method (CVM).

%
\subsection{2-D Cluster Variation Method Overview }
\label{subsecCVM-:overview}
%

This Technical Report introduces how a CORTECON(R) can be used to construct the set of representational units $(r)$. We envision the formulation of a representational system whose component elements are pre-specified, and which is distinct from the external system that is being represented. Further, we envision a total system (external together with representation) in which both the external and representational systems can, and indeed do, separately achieve free energy minimization. Their ability to do this requires, of course, that a \textbf{\textit{free energy equation exists for each of these respective systems}}. 

One way in which we can have a system that allows for \textit{both} free energy minimization and suitable modeling richness is to use a 2-D system constructed as a grid of bistate nodes, as was shown in the previous Fig.~\ref{fig:computational-engine-1}}. In such a system, we can use the cluster variation method (CVM) to compute a free energy, for which the entropy term is more complex than is typically used. The theoretical basis for this was first developed by Kikuchi \cite{Kikuchi_1951_Theory-coop-phenomena}, and then jointly by Kikuchi and Brush \cite{Kikuchi-Brush_1967_Improv-CVM}. A more recent description is provided in Maren \cite{Maren_2016_CVM-primer-neurosci}. 

The key measurable variables within a CVM system are the \textit{configuration variables}. In addition to the simple identification of proportional numbers of ``on'' (\textbf{A}, black) and ``off'' (\textbf{B}, white) nodes, these configuration variables also account for the various kinds of nearest-neighbor ($y_i$) and next-nearest-neighbor ($w_i$)  pairs, as well as the six different kinds of triplets (denoted $z_i$). (A figure in Appendix~\ref{sec:Appendix-cluster-variation-method} illustrates these different configuration variables.)

In using the CVM method for describing the free energy, the equilibrium distribution of nodes is governed by two enthalpy parameters, $\varepsilon_0$ and $\varepsilon_1$. These parameters are the only ``tunable'' parameters available in the CVM formulation, and thus are identified with $\theta$, as used by Friston (op. cit.) and Beal \cite{Beal_2003_Variational-algorithm-approx-Bayes-inference}.  

For the case where the distribution of units into the two states \textbf{A} and \textbf{B} is equiprobable, the activation enthalpy parameter $\varepsilon_0$ is by definition zero. This leaves only a single ``tunable''  parameter, $\varepsilon_1$; the interaction enthalpy parameter. For this specific case, where the fractions of \textbf{A} and \textbf{B} nodes are equal, there is an analytic solution that provides the relative equilibrium fractions of the different configuration variables as a function of $\varepsilon_1$. Due to how the equilibrium solution for the configuration variables is expressed, it is easier to refer to a parameter that is a function of $\varepsilon_1$ than that specific value itself. Thus, we normally use the \textit{h-value}, where $h=exp(2\varepsilon_1)$. 

The analytic solution is not particularly accurate for larger \textit{h-values} (e.g., $h > 1.6$), but it provides a starting point for computational free energy minimization, for a given system and a corresponding given \textit{h-value}. 

Formally speaking, to apply the free energy principle (or indeed variational Bayes to any given data), it is entirely sufficient to specify a generative model in terms of a joint distribution over data and their latent causes or, in a Markov blanket partition, sensory and external states (where sensory states are augmented with internal and active states in the Markov blanket formalism). 

Appendix C describes the implicit generative model that CVM entails – to give an idea of the sort of data it can generate – and therefore explain or recognize. The following subsections present an illustration of how the 2-D CVM would actually look in application.

%
\subsection{CVM Illustration }
\label{subsec:CVM-illustration}
%

The following Figure~\ref{fig:CVM-2D_Scale-free_1024-and-scale-free-256-nodes_2019-06-18_full_first-fig_crppd} provides a conceptual  illustration for two systems;  (a) corresponds to the external ($\tilde{\psi}$) system, and (b) corresponds to the representational system  ($\tilde{r}$). (This follows notation introduced by Friston \cite{Friston-et-al_2015_Knowing-ones-place-free-energy-pattern-recognition}.) The grid pattern of active (\textbf{A}) and inactive (\textbf{B}) units is suitable for modeling using a 2-D cluster variation method (CVM) free energy equation, as described in Maren \cite{Maren_2016_CVM-primer-neurosci, AJMaren-2021-2D-CVM-Topography}. 

Neither of the two systems depicted in Fig.~\ref{fig:CVM-2D_Scale-free_1024-and-scale-free-256-nodes_2019-06-18_full_first-fig_crppd} are at equilibrium; both are hand-crafted with the intent of embodying a \textit{scale-free} type of system. 

Each system was constructed to have an equal number of nodes in states \textbf{A} and \textbf{B}, or ``on'' and ``off'' states. This implies that the activation enthalpy parameter $\varepsilon_0 = 0$. However, the likely value for the interaction enthalpy parameter $\varepsilon_1$ was unknown for each of the two systems. (We typically work with the \textit{h-value} instead of with $\varepsilon_1$, where $h=exp(2\varepsilon_1)$.) 

Thus, our first goal  was to identify a likely \textit{h-value} candidate for each of the two systems. Pragmatically, we focused on the external, or $\tilde{\Psi}$, system, as it was more extensive and allowed for a richer set of at-equilibrium patterns to evolve. Our second goal was then to computationally bring that system into a free energy minimum for that particular \textit{h-value}, which was determined to be approximately 1.2. (Detailed experimental results are in Maren \cite{AJMaren_2019_Expt-Results_Two-epsilon-params}.)

In actuality, we would most likely not know the \textit{h-values} for the external $\tilde{\Psi}$ system. We use the plural ``\textit{h-values}'' instead of the singular, because even if we had an equiprobability of \textbf{A} and \textbf{B} nodes, we would not necessarily have a single \textit{h-value} that would characterize the system. 

This is because the distribution of local configuration variables - that is, the nearest-neighbor, next-nearest-neighbor, and triplet configurations - would not likely correspond to an equilibrium state. Instead, each distinct configuration variable would have a specific \textit{h-value} that would correspond with it, and there would be a range of \textit{h-values} associated with a given system. 

We envision how this would be considered in the active inference context, where we have an external system ($\tilde{\psi}$), and are seeking to represent it with an internal system of representation units  ($\tilde{r}$). We would likely be able to sample the configuration variables at various locations, and for various degrees of granularity, for the external system. These would become the inputs ($\tilde{s}$) to the units in the representational system $\tilde{r}$. 

The two systems shown in Figure~\ref{fig:CVM-2D_Scale-free_1024-and-scale-free-256-nodes_2019-06-18_full_first-fig_crppd} have similar pattern configurations, with the exception that the larger system shown in (a) has clusters that are proportionately larger than the clusters in the smaller-scale representational system (b). The natures of the patterns within each, though, are much the same. Thus, to a first order, they should have similar (reduced) free energies. More to the point, we envision that the representational system shown in (b) can be brought into alignment with that of (a), or (more specifically) be brought to a free energy minimium with the same (or similar) \textit{h-value} as with the \textit{h-value} corresponding to the external system of (a).

We note, of course, that since neither system is likely to be at equilibrium, that initially we will not have a single  \textit{h-value} for either. One task is to find an  \textit{h-value} that provides a ``best fit'' to each of the systems. To that end, we have devised a new divergence measure, the \textit{(reverse) Kikuchi-Maren divergence}, which is conceptually akin to the (reverse) Kullback-Leibler divergence \cite{AJMaren-2022-Variational-Approach-2D-CVM, Maren_2024_Minding-Your-Ps-and-Qs-Kullback-Leibler-Divergence}. 

\begin{figure}[htbp]
    \centering
        \includegraphics[trim=0cm 0cm 0cm 0cm, clip=true,  width=0.90\textwidth]{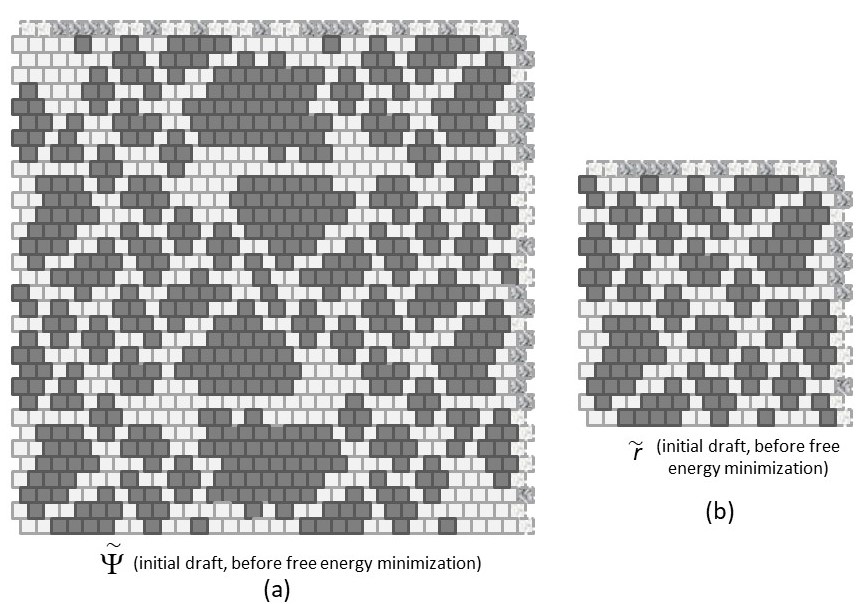}
   \caption{Illustration of two systems, arranged so that a 2-D CVM-based free energy can be directly computed for each. (a) The external system, with units denoted $\tilde{\psi}$. (b) The representational system, showing only the representational units,($\tilde{r}$). The Markov blanket around the grid of representational units is not shown in this figure. The dark and light-shaded grey and mottled units to the upper and right edges of each system illustrate the wrap-around from the left and bottom edges, used to compute the configuration variables leading to the free energies of each system. Both systems show an approximate scale-free distribution of islands of dark (\textbf{A}) units in a sea of white (\textbf{B}) units. The systems are designed with equiprobable distribution of units into states \textbf{A} and \textbf{B} ($x_A = x_B = 0.5$), so that the (reduced) free energies of each can be computed directly, using the analytic solution provided in Maren \cite{Maren_2016_CVM-primer-neurosci, AJMaren-2021-2D-CVM-Topography}. Details of the corresponding thermodynamic calculations are found in Maren \cite{AJMaren_2019_Expt-Results_Two-epsilon-params}. The systems shown in this figure have been hand-designed to illustrate a potential scale-free configuration; they have not yet been brought into free energy minimization.} 
    \label{fig:CVM-2D_Scale-free_1024-and-scale-free-256-nodes_2019-06-18_full_first-fig_crppd}
\end{figure}

%
\subsection{Interpreting the CVM in the Variational Bayes Framework}
\label{subsec:intepreting-CVM}
%

\begin{figure}[htbp]
    \centering
        \includegraphics[trim=0cm 0cm 0cm 0cm, clip=true,  width=1.00\textwidth]{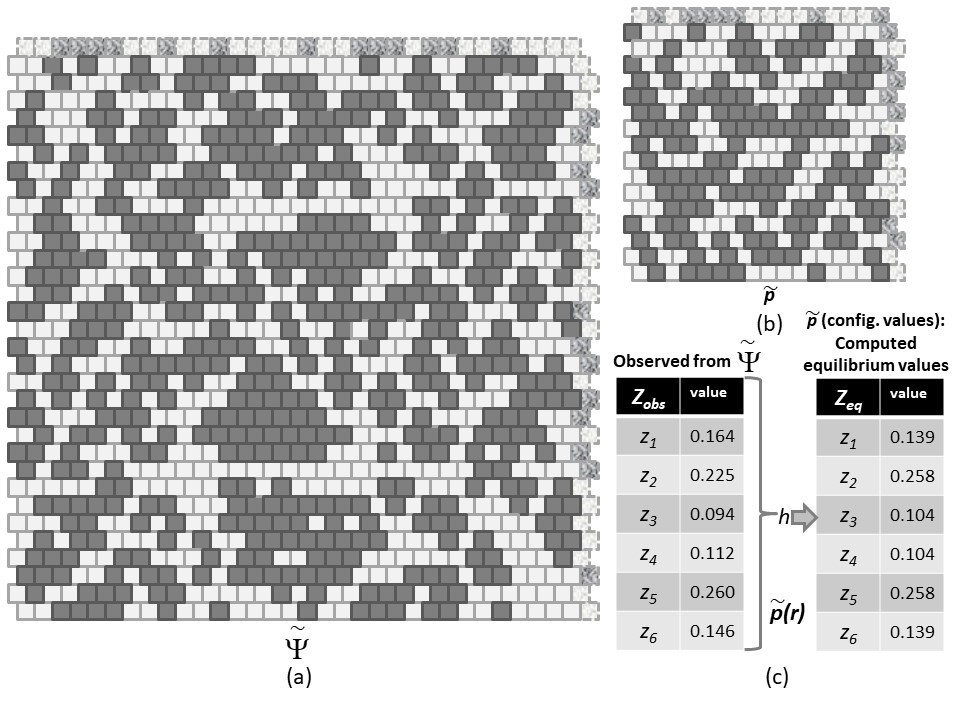}
   \caption{(a) The external system $\tilde{\Psi}$ has been brought to a free energy minimum for the case where $h = 1.2$.  Sampling this system provides different inputs to the representational system with units $\tilde{r}$. In reality, we would not directly know the \textit{h-values} corresponding to $\tilde{\Psi}$. However, we would trust that the system $\tilde{r}$, taking its configuration values from sensing applied to the units $\tilde{\psi}$, would also be at equilibrium. Finding the \textit{h-values} for $\tilde{r}$  would give us the parameters for the  model $\tilde{p}$, shown in (b). In this particular case, as the full set of \textit{h-values} corresponding to different configuration values still needs to be developed, the system $\tilde{p}$ was devised for illustration purposes  by performing free energy minimization on $\tilde{r}$, shown in the previous Fig.~\ref{fig:CVM-2D_Scale-free_1024-and-scale-free-256-nodes_2019-06-18_full_first-fig_crppd}, for $h=1.2$. The equilibrium configuration values for $\tilde{\Psi}$ and $\tilde{p}$ are shown in (c).} 
    \label{fig:CVM-2D_Scale-free_1024-and-scale-free-256-nodes_2022-11-03_full_2nd-fig_crppdl}
\end{figure}

The variational Bayes method provides a framework for a new computational engine, and the first step towards this  is illustrated in Figure~\ref{fig:CVM-2D_Scale-free_1024-and-scale-free-256-nodes_2022-11-03_full_2nd-fig_crppdl}.

As with the previous Figure~\ref{fig:CVM-2D_Scale-free_1024-and-scale-free-256-nodes_2019-06-18_full_first-fig_crppd}, the larger-scale system on the left (a) corresponds to the external system. In the case of Figure~\ref{fig:CVM-2D_Scale-free_1024-and-scale-free-256-nodes_2022-11-03_full_2nd-fig_crppdl}, though, the system has been brought to a free energy minimum, for the case where $\varepsilon_0 = 0$ (since the system has been designed with equiprobable distribution of \textbf{A} and \textbf{B} nodes), and where $h = 1.2$. (Recall that $h = exp(2\varepsilon_1)$.) The selection of $h = 1.2$ was done by first (computationally) counting the distribution of all the different nearest-neighbor and next-nearest-neighbor pairs, as well as the different triplets, and then computing their relative fractions as configuration variables. 

Using the analytic solution for equilibrium values of the configuration values as functions of \textit{h}, it was possible to estimate a range of possible values for \textit{h}. (Since the original system of Figure~\ref{fig:CVM-2D_Scale-free_1024-and-scale-free-256-nodes_2019-06-18_full_first-fig_crppd} was not at equilibrium, the various configuration values corresponded to different analytic \textit{h-values}.) For simplicity, the next step was done using $h = 1.2$, which was the \textit{h-value} corresponding to the nearest-neighbor pairs for unlike nodes. For details, see Maren \cite{AJMaren_2019_Expt-Results_Two-epsilon-params}.) 

We can see, in Figure~\ref{fig:CVM-2D_Scale-free_1024-and-scale-free-256-nodes_2022-11-03_full_2nd-fig_crppdl}, that in both the external system (a) and the representational system (b), some of the respective units $\tilde{\psi}$ as well as the units in the representation $\tilde{p}$ have taken on different values. (The total numbers of units in states \textbf{A} and \textbf{B} remains equal, in each of these systems. This allows us to apply the analytic solution as a starting point for selecting an \textit{h-value}.)

For each of these systems, we see a set of configuration variables that now represent at-equilibrium values for the case where $h=1.2$. Under normal circumstances, we would not be directly bringing the external system $\tilde{\Psi}$ to equilibrium; we would instead be sampling it with our sensory units $\tilde{s}$. These sensory units would influence the representational units $\tilde{r}$. The representational system could then, following the precepts of Action Perception Divergence \cite{Hafner-et-al_2022_Action-perception-divergence}, direct action agents $\tilde{a}$ to influence the external system  $\tilde{\Psi}$.

As we bring the model of the representational system into free equilibrium, the configuration variables reflect what would be the case for overall equilibrium in the external system $\tilde{\Psi}$. These configuration variables describe the topography of an at-equilibrium system. Further, the equilibrium state for this set of configuration variables corresponds to a specific \textit{h-value}, which here functions as a model parameter $\theta$. 

In a more complete build-out of this approach, we would be able to vary both the activation enthalpy $\varepsilon_0$ and the interaction enthalpy $\varepsilon_1$ parameters, where for our purposes, $h=exp(2\varepsilon_1)$. This means that we would be able to identify a full set of configuration variables with just the parameter set $\theta = (\varepsilon_0, \varepsilon_1)$.  

As the phase space map for various equilibrium configuration values versus different \textit{h-values} becomes worked out, it will be possible to find a corresponding set of \textit{h-values} given an initial set of configuration values. It will then be possible to perform free energy minimization on the system $\tilde{r}$, using different candidate \textit{h-values} together with estimates for the activation enthalpy parameter $\varepsilon_0$, to obtain a resultant model $\tilde{q}$ that provides an acceptable fit to the units in the representational system $\tilde{r}$. 

This mechanism can further be used to model the external system $\tilde{\Psi}$ as it moves through different states, with various corresponding \textit{h-values}. This means that we would potentially have a means for modeling evolving system dynamics over time. Moreover, the model would be encapsulated into an \textit{h-values} trajectory, which would be a relatively simple $\theta$ model.  

\vspace{5 mm}

%
\section{Conclusions}
\label{sec:Conclusions}
%

\begin{flushright}
``\textit{`What?' Female} was an alien language, but he usually could translate it well enough to understand what was being said. But this [was] ...''\\

\textit{Tangled Webs}\\

Anne Bishop (2008), p. 159 (Hardcover edition). 
 \end{flushright}

This Technical Report has served three purposes:

\begin{enumerate}
\item Perform a ``Rosetta Stone'' translation,
\item Describe how the external and representational systems begin as separate entities, each of which can (separately) come to free energy equilibrium, and
\item Introduce a method for system representation that could, within itself, undergo free energy minimization in order to yield a resulting model which could be described using only one or two parameters (the $\theta$ elements of the model \textit{q}). 
\end{enumerate}

There has been substantial grumbling within the research community about how difficult it has been to read and understand Karl Friston's various articles. (See Freed \cite{Freed_2014_Research-Digest-Commentary-Friston} as just one example.) Some of this is notational; Friston has changed his notation subtly - just enough to be difficult for (and perhaps maddening to) the reader - throughout his various articles. Yet, a growing sense that he's presenting a very useful approach is driving more and more researchers to attempt to read his works. 

This desire to understand the fundamental variational Bayes approach, and Friston's extension to describing external systems and their corresponding representational systems (separated by a Markov blanket), is growing. The variational Bayes methods are receiving greater attention, as a next step in machine learning methods, as described by Yellapragada and Konkimalla \cite{Yellapragada-and-Konkimalla_2019_Variational-Bayes}. Wainwright and Jordan (2008) have published an extensive tutorial on variational Bayes, setting it in the overarching context of graph theory \cite{Wainwright-and-Jordan_2008_Graph-models-exp-fam-var-inf}.

Thus, the first intention of this work has been to perform a ``Rosetta Stone'' translation between the variational Bayes derivation as given by Beal  \cite{Beal_2003_Variational-algorithm-approx-Bayes-inference} and the subsequent ones given by Friston \cite{Friston_2010_Free-energy-principle-unified-brain-theory, Friston_2013_Life-as-we-know-it, Friston-et-al_2015_Knowing-ones-place-free-energy-pattern-recognition}, with particular attention to how the shifts in notation reflect a move from an envisioning where both the external and representational systems are predicated on the same underlying variables $x_i$ to one in which the external and model states can be separated. This should make Friston's works more readable to the broader scientific community. 

The second intention has been envisioning how this formulation can be used for a scenario in which the external and representational systems begin as separate entities, each of which can (separately) come to free energy equilibrium. (As an example, see Friston and  Frith \cite{Friston_and_Frith_2015_Duet-for-one} for an example cast in terms of communication and birdsong.) This is important, because we have not typically thought about how various systems (whether external or representational) need to be expressed in a way that allows free energy minimization. 

As one of Friston's key points is that free energy minimization underlies crucial processes (including brain processes), we need to have a better understanding of this premise. Some work, such as that done by a team led by Moran and published by Cullen et al. \cite{Cullen-et-al_2018_Active-inference}, has already shown the validity of this approach. That work shows how a variational Bayes approach can outperform reinforcement learning, within a specific and constrained operational environment. Those experiments indicate a promising direction for future investigation. 

The third intention has been to introduce a means for representing a system, that is, the $\tilde{r}$ component of the system model, using a 2-D cluster variation method approach. This gives us a representation for which we can write a free energy equation, and thus carry out explicit free energy minimization, leading to parameter identification for the free energy-minimized state. This approach has only been briefly sketched; greater expostulation is provided elsewhere and further developmental work is underway. 

It is possibly this last intention that will prove the most valuable over time. There is currently a paucity of useful models for which free energy minimization is an inherently appropriate method. Specifically, the well-known Ising model (which has become dearly beloved within deep learning circles) does not offer sufficient richness for more complex system modeling. The 2-D CVM approach allows both for richness in expression and a simplicity in terms of the two parameters that govern this expression. 

The impediment thus far has been that the CVM approach has been theoretically obscure, and its practical capabilities so far unknown. In fact, the phase space behavior of this model - in terms of identifying how the activation and interaction enthalpy parameters impact the resulting free energy-minimized states - has not yet been fully mapped out. This is largely a computational problem, somewhat aided and abetted by (limited) analytic solutions. Work on the 2-D CVM is underway, which should make this model available for use in the near future as a means for implementing model systems that can, within their own nature, be free energy-minimized. This will introduce a new kind of modeling capability for a wide range of applications.

\vspace{6pt}


\section*{Acknowledgements}

I am enormously indebted to Karl Friston for careful, detailed, and thoughtful reviews, together with very useful suggestions for rewording a few explanations.


\section*{Declaration of No Conflicts}

The author declares that the research was conducted in the absence of any commercial or financial relationships that could be construed as a potential conflict of interest.


\section*{Code Availability}

The initial Python code for computing the 2-D CVM configurations, together with the corresponding entropy, enthalpy, and free energy values, is available from the author in the \textit{ajmaren} GitHub repositories. See Maren (2018) \cite{AJMaren-TR2018-001v2-V-and-V} for the code verification and validation documentation.

More recently, there has been an effort to transition the original Python code to an object-oriented Python code set. This object-oriented code is available at tje Themesis GitHub repository \cite{Themesis-GitHub}. This code is made available under the MIT License Agreement. Themesis has developed an extensive set of YouTube code walkthrough tutorials \cite{Themesis-CORTECON-YouTube-Playlist}.

Those who ``Opt-In'' with Themesis (www.themesis.com/themesis/) will receive word when code is released, along with word on new experimental results with the 2-D cluster variation method. 


\appendix

%
\section{Appendix: Fundamental Thermodynamic Concepts}
\label{sec:Fund-thermo-concepts}
%

\renewcommand{\theequation}{A-\arabic{equation}}
\setcounter{equation}{0}  

In various commentaries, researchers note that the term ``thermodynamic free energy,'' as used by Friston (op. cit.) does not really correspond to a a true thermodynamic free energy. Similarly, there is a difference in the enthalpy term, as computed and used in Friston's work (and in others using the variational Bayes method) and in the notion of enthalpy as it is found in statistical thermodynamics. 

This Appendix briefly overviews some of the key concepts in statistical (and classical) thermodynamics, so that it is easier to compare the formalisms resulting from the variational Bayes method described in the body of this Report with the corresponding formalisms from statistical thermodynamics, as is commonly known. 

Note that the thermodynamic quantities of free energy ($F$), enthalpy ($H$), and entropy ($S$) are all extensive variables; their values are subject to the number of total units in a given system. However, we will work throughout with the \textit{reduced variables}, ($\bar{F}$), the enthalpy ($\bar{H}$), and the entropy ($\bar{S}$), for which the previous values have been divided through by $Nk_{\beta}T$, where $N$ is the total number of units in the system, $k_{\beta}$ is Boltzmann's constant, and $T$ is the temperature. This reduces all values to dimensionless quantities which are independent of the size of the systems under consideration. For the remainder of this work, the overhead bar notation on the reduced thermodynamic variables will be dropped; all quantities are understood to be reduced. 

The well-known thermodynamic equation for the free energy is given as

\begin{equation}
\label{eqn:basic-eqn-free-energy-classic}
  \begin{aligned}
    F =  H \: - \:  S,    
  \end{aligned}
\end{equation} 

\noindent
where $F$, called the \textit{free energy}, is the energy available to do work, $H$ is the enthalpy (also called the internal energy) of a system, and $S$ is the entropy. In this \textit{reduced} free energy equation, we have already divided through by the (absolute) temperature $T$.

From a statistical thermodynamics perspective, in order to calculate any of these terms, we first need the probabilistic distribution of units in the system among available states. (Several of the following equations are drawn from Maren \cite{AJMaren-2013-Stat-Mech-Basic-Theory, AJMaren-BOOK-Stat-Mech-NN-AI-Precis}.)

%
\subsection{The Classic Partition Function}
\label{subsec:Classic-partition-function}
%

To introduce the statistical thermodynamics approach to describing the free energy of a system, we begin with a simple (and classical) expression for the probability of finding a system in the quantum state $i$, characterized by energy $E_i$, as

\begin{equation}
\label{eqn:basic-probabilty-of-states-proportional}
  \begin{aligned}
    P_i \propto \exp(-E_i/{k_{\beta}T}),     
  \end{aligned}
\end{equation}  

\noindent
where $k_\beta$ is Boltzmann's constant and $T$ is temperature. 

We note that the energy $E_i$ describes the energy of the entire system of $N$ units, and that there very well may be degeneracy in the ways in which various units can be assembled to yield a certain energy. 

As is true for any probability distribution,

\begin{equation}
\label{eqn:probabilty-dist-equals-1_v3}
  \begin{aligned}
    \sum_{i=1}^N P_i = 1.     
  \end{aligned}
\end{equation}

In Eqn.~\ref{eqn:basic-probabilty-of-states-proportional}, we stated a proportionality. Now, to find the proportional constant $c$, we state that

\begin{equation}
\label{eqn:basic-probabilty-of-states-equality}
  \begin{aligned}
    P_i = c \exp(-E_i/{k_{\beta}T}),     
  \end{aligned}
\end{equation}  

\noindent
which gives us 

\begin{equation}
\label{eqn:probabilty-dist-equals-1_v2}
  \begin{aligned}
    \sum_{i=1}^N P_i =  \sum_{i=1}^N c \exp(-E_i/{k_{\beta}T}) = c \sum_{i=1}^N  \exp(-E_i/{k_{\beta}T}) = 1,     
  \end{aligned}
\end{equation}

\noindent
so that 

\begin{equation}
\label{eqn:proportional-const-for-probabilty-dist}
  \begin{aligned}
    c = 1 / \sum_{i=1}^N  \exp(-E_i/{k_{\beta}T}).     
  \end{aligned}
\end{equation}

This sum of probabilities becomes a valuable quantity in and of itself; we refer to it as the \textit{partition function}, $Q$ (and in some sources referred to as $Z$), so that

\begin{equation}
\label{eqn:def-Q}
  \begin{aligned}
    Q = \sum_{i=1}^N  \exp(-E_i/{k_{\beta}T}).     
  \end{aligned},
\end{equation}

\noindent
which allows us to phrase a distinct probability as

\begin{equation}
\label{eqn:probability-with-Q}
  \begin{aligned}
    P = \exp(-E_i/{k_{\beta}T}) / Q.     
  \end{aligned}
\end{equation}

%
\subsection{The Classic Enthalpy Formulation}
\label{subsec:Classic-enthalpy-formulation}
%

Now, we define the average energy $H$, or \textit{enthalpy}, of a system to be the expectation for the energy of the system, which can be described as the average of the sum of all the energies of the system,

\begin{equation}
\label{eqn:simple-H}
  \begin{aligned}
    H &= \langle\langle E_i \rangle\rangle \\
    &= \sum_{i=1}^N E_i  P_i \\
    &= \sum_{i=1}^N E_i \exp(-E_i/{k_{\beta}T}) / Q \\
    &= \frac {1} {Q} \sum_{i=1}^N E_i \exp(-E_i/{k_{\beta}T}).         
  \end{aligned}
\end{equation} 

For simplicity, we introduce the notation that $\beta = 1/{k_{\beta}T}$, so that

\begin{equation}
\label{eqn:simple-H-2}
  \begin{aligned}
    H = \frac {1} {Q} \sum_{i=1}^N E_i \exp(-{\beta} E_i).         
  \end{aligned}
\end{equation}

Now, we notice that we can make a further simplification, by observing that

\begin{equation}
\label{eqn:deriv-energy-times-prob-term}
  \begin{aligned}
    E_i \frac {\exp(-{\beta} E_i)} {Q} = - \frac {1}{Q}\frac {\partial} {\partial \beta} \exp(-{\beta} E_i)         
  \end{aligned}
\end{equation}

\noindent
so that

\begin{equation}
\label{eqn:deriv-energy-times-prob-term-2}
  \begin{aligned}
    E_i \exp(-{\beta} E_i) = - \frac {\partial} {\partial \beta} \exp(-{\beta} E_i)         
  \end{aligned}
\end{equation}

Thus, we can rewrite Eqn.~\ref{eqn:simple-H-2} as 

\begin{equation}
\label{eqn:simple-H-3}
  \begin{aligned}
    H &=  \sum_{i=1}^N \frac {E_i \exp(-{\beta} E_i)} {Q} \\
&= - \frac {1} {Q} \sum_{i=1}^N \frac {\partial} {\partial \beta} \exp(-{\beta} E_i) \\
&= - \frac {1} {Q} \frac {\partial} {\partial \beta} \sum_{i=1}^N  \exp(-{\beta} E_i).         
  \end{aligned}
\end{equation}

We notice, from Eqn.~\ref{eqn:def-Q} that the entire of the summation in Eqn.~\ref{eqn:simple-H-3} is exactly $Q$ itself. Thus, we can rewrite the expression for $H$ in Eqn.~\ref{eqn:simple-H-3} as 

\begin{equation}
\label{eqn:H-in-terms-of-Q}
  \begin{aligned}
    H &= - \frac {1}{Q} \frac {\partial} {\partial \beta} \sum_{i=1}^N  \exp(-{\beta} E_i)
&= - \frac {1}{Q}  \frac {\partial} {\partial \beta}     Q.       
  \end{aligned}
\end{equation} 

We recall the expression for the derivative of a logarithmic function, that 

\begin{equation}
\label{eqn:def-deriv-log-y}
  \begin{aligned}
    \frac {\partial} {\partial x} ln(y(x)) =   \frac {1}{y} \frac {\partial y(x)} {\partial x}.       
  \end{aligned}
\end{equation} 

Thus, noticing the correspondence between $Q = Q(\beta) = y(x)$, we can identify that

\begin{equation}
\label{eqn:deriv-log-Q}
  \begin{aligned}
    \frac {\partial} {\partial \beta} ln(Q(\beta)) =   \frac {1}{Q} \frac {\partial Q(\beta)} {\partial \beta}.       
  \end{aligned}
\end{equation} 

This allows us to rewrite Eqn.~\ref{eqn:H-in-terms-of-Q} as

\begin{equation}
\label{eqn:H-in-terms-of-ln-Q}
  \begin{aligned}
    H &= - \frac {1}{Q}  \frac {\partial} {\partial \beta}     Q
&= - \frac {\partial} {\partial \beta} ln(Q(\beta)) 
&= - \frac {\partial} {\partial \beta} ln(Q),       
  \end{aligned}
\end{equation} 

\noindent
dropping the functional dependence of $Q$ on $\beta$ in the last expression. 

This gives us a very powerful means for expressing the enthalpy, or internal energy, of a system in terms of $ln(Q)$. However, it is not a completely general expression for $H$. Instead, we note that by taking the (negative of) the derivative of $ln(Q)$ with respect to $\beta$, we are moving to a value of $ln(Q)$ that maximizes (minimizes) the function in terms of temperature (actually, $1/T$). This then leads us to the value of enthalpy, $H$, that helps to minimize the overall value for the free energy, $F$. The value of $H$ that would occur in this case (taking into account the impact of entropy), would then be the expected value of $H$ when the system is at equilibrium. 

What we can generalize from this is not so much the specific formulation for $H$ (as a partial derivative of $\ln(Q)$ with respect to $\beta$), but rather, that we are looking for an expected value of $H$. Depending on how the system is formulated, it could be a different resulting expression.

%
\section{Appendix B: Variational Free Energy: Enthalpy and Entropy}
\label{sec:Appendix-var-FE-enthalpy-entropy}
%

  \renewcommand{\theequation}{B-\arabic{equation}}
  \setcounter{equation}{0}  

This Appendix provides more detail on the derivation of the free energy equation as given in Friston \cite{Friston_2013_Life-as-we-know-it, Friston-et-al_2015_Knowing-ones-place-free-energy-pattern-recognition}, using the formulation established by Beal \cite{Beal_2003_Variational-algorithm-approx-Bayes-inference}, and making the correspondence between the notations of each. It yields the expression of the variational free energy in terms of what Friston refers to as expected energy or enthalpy (and in one source \cite{Friston-et-al_2015_Knowing-ones-place-free-energy-pattern-recognition}, as ``thermodynamic free energy'') plus an entropy term. This expected energy or enthalpy is actually the expectation for the (negative of the) log-likelihood of a certain distribution, as given in Eqn.~\ref{eqn:L-thermodynamic-free-energy-summation_joint-probability}, which is restated here as

\begin{equation}
\label{eqn:L-thermodynamic-free-energy-summation_joint-probability-in-Appendix-B}
  \begin{aligned}
    L(\tilde{x}) = L(\tilde{{\psi}},\tilde{s},\tilde{a},\tilde{r})
 = -\ln {p(\tilde{{\psi}},\tilde{s},\tilde{a},\tilde{r})},   
  \end{aligned} \nonumber
\end{equation}

If we were to use a 2-D CVM grid to represent the external and internal (representational) states, we would take our probabilities as the actual fractional values for the different configuration variables, and we would use the entropy term from the 2-D CVM formalism (see Appendix C, and references cited therein) to suggest a structuring for the probabilities in the $L(\tilde{x})$ and entropy terms.

We will, very shortly, follow a line of argument introduced by Beal. One of the steps that he makes at a certain conclusion gives a form of the enthalpy equation, which is fundamentally the same as used by Friston (op. cit.), and is the one that we have shown in Section~\ref{sec:var-FE-log-likelihood-expectation-and-entropy}. It will be that

\begin{equation}
\label{eqn:basic-free-energy-Beal-H-only}
  \begin{aligned}
    H  =  \sum_{i=1} \int dx_i \: q_{x_i}(x_i) \: ln  \: p(x_i,y_i|\theta).    
  \end{aligned}
\end{equation} 

In the next few paragraphs, we will show that Eqn.~\ref{eqn:basic-free-energy-Beal-H-only} serves analogously to Eqn.~\ref{eqn:H-in-terms-of-ln-Q}. It fits the role of enthalpy in what is being described in a ``free energy'' formalism. However, it is not a derivative equation; it is the expectation for the natural logarithm of the probability of certain variables. 

By examining the correspondence between the two expressions (the variational Bayes in comparison with the thermodynamic), when we come to expressing the full free energy for the representational system (using the formulation expressed by Beal), the free energy derivation will be much more lucid. 

Beal \cite{Beal_2003_Variational-algorithm-approx-Bayes-inference} (Sect. 2.2.1) introduces the scenario  for parameter learning with the notion of a generative model with hidden variables \textit{x} and observed variables \textit{y}, where the dataset of observed variables $y = \{y_1,...,y_n\}$ are generated by the set of hidden variables $x = \{x_1,...,x_n\}$, and where the \textit{n} items in each case are independent and identically distributed (i.i.d.). He identifies the parameters describing the (potentially) stochastic dependencies between variables as $\theta$. The probability distribution for observing $y$ (Beal, 2003, Eqn. 2.9) is given as

\begin{equation}
\label{eqn:basic-likelihood-integral-y-x-and-theta}
  \begin{aligned}
    p(y|\theta) = \prod_{i=1}^n p(y_i|\theta)  = \prod_{i=1}^n  \int dx_i p(x_i,y_i|\theta).    
  \end{aligned}
\end{equation}

This tells us that the probability distribution of the observable variables $y_i$ is conditioned on the parameters $\theta$. 

We compare this with our earlier expression for the probability that a system would be in the $i^{th}$ energy state, where we stated in Eqn.~\ref{eqn:def-Q} that the partition function $Q$ (which served as a normalizing factor) is given as

\begin{equation}
\label{eqn:def-Q-restated}
  \begin{aligned}
    Q = \sum_{i=1}^N  \exp(-E_i/{k_{\beta}T}), \nonumber     
  \end{aligned}
\end{equation}

\noindent
and from Eqn.~\ref{eqn:probability-with-Q} that probability is given as

\begin{equation}
\label{eqn:probability-with-Q-restated}
  \begin{aligned}
    P = P_i = \exp(-E_i/{k_{\beta}T}) / Q. \nonumber     
  \end{aligned} 
\end{equation}

The difference in the two formulations is that, in Eqn.~\ref{eqn:probability-with-Q}, we are dealing with the probability of finding a system in a given $i^{th}$ energy state, where there could indeed be multiplicity of units inhabiting various components of that overall system's energy state, and there can also be degeneracy in the units inhabiting the various states. This means, various units, indistinguishable from each other, can inhabit the energy states so that there are multiple means of counting the units in various states and coming up with the same overall $i^{th}$ system energy. 

In contrast, the formulation given in Eqn.~\ref{eqn:basic-likelihood-integral-y-x-and-theta} deals with a full \textbf{\textit{set}} of observed variables, $y = \{y_1,...,y_n\}$.

As we know from the formulation of joint probabilities, the joint probability of observing variable $y_1$ in a given state $i$, $p(y_1=i)$, together with the probability of observing variable $y_2$ in its own given state $j$, $p(y_2=j)$, is $p(y_1=i)p(y_2=j)$. Thus, for a set of such states, $y = \{y_1,...,y_n\}$, we need the product of each of the unique variable's probability, which yields the first part of Eqn.~\ref{eqn:basic-likelihood-integral-y-x-and-theta};

\begin{equation}
\label{eqn:basic-likelihood-integral-y-only-on-theta}
  \begin{aligned}
    p(y|\theta) = \prod_{i=1}^n p(y_i|\theta).  
  \end{aligned}
\end{equation}

As a second step in making the correlation between Eqn.~\ref{eqn:basic-likelihood-integral-y-x-and-theta} and the combination of Eqns.~\ref{eqn:def-Q} and \ref{eqn:probability-with-Q}, we examine the formulation (from Eqn.~\ref{eqn:basic-likelihood-integral-y-x-and-theta})

\begin{equation}
\label{eqn:dependence-of-y-on-x-and-theta}
  \begin{aligned}
    p(y_i|\theta)  = \int dx_i p(x_i,y_i|\theta).    
  \end{aligned}
\end{equation}

We note that Eqn.~\ref{eqn:dependence-of-y-on-x-and-theta} essentially states that the observable variable(s) $y_i$ are dependent on a hidden, unobservable set of variables $x_i$. For each distinct value (or set of values) $y_i$, there are potentially multiple contributions, drawn over the complete set of   $x_i$. The notation here (keeping to that used by Beal) is just a little ambiguous; it does not mean that the set of countable variables $x_j$ is exactly the same as the set of countable variables $y_i$. Rather, it is saying that for a given instance of a set of variables $y_i$, there is a corresponding set of variables $x_i$, $x_i \equiv \lbrace X_i \rbrace = \lbrace x_{i,1}, x_{i,2},...,x_{i,J} \rbrace $, where $J$ is the total number of elements in the set of variables $\lbrace X_i \rbrace$ contributing to values for $y_i$.

This means, since each of these \textit{contribute} to the value for $y_i$, that summation (rather than multiplication) is needed. Since the realm of variables constituting $\lbrace X_i \rbrace$ is assumed here to be large, integration rather than summation is used. This, then, gives us a plausibility argument in support of Eqn.~\ref{eqn:basic-likelihood-integral-y-x-and-theta}. 

Before moving on, we need one more step. Eqn.~\ref{eqn:basic-free-energy-Beal-H-only} involves more than the probability distribution of Eqn.~\ref{eqn:basic-likelihood-integral-y-x-and-theta}. We note that in Eqn.~\ref{eqn:basic-free-energy-Beal-H-only}, we have a summation rather than a product, the natural logarithm of $p(x_i,y_i|\theta)$ as opposed to just a simple statement of the probability, and an additional modulating term, $q_{x_i}(x_i)$. 

To see how these occur, we go back to our original definition of the enthalpy, $H$, as given in Eqn.~\ref{eqn:H-in-terms-of-ln-Q};

\begin{equation}
\label{eqn:H-in-terms-of-ln-Q-repeat}
  \begin{aligned}
    H &= - \frac {1}{Q}  \frac {\partial} {\partial \beta}     Q
&= - \frac {\partial} {\partial \beta} ln(Q(\beta)) 
&= - \frac {\partial} {\partial \beta} ln(Q). \nonumber      
  \end{aligned}
\end{equation} 

Suppose that we have a probability distribution given as

\begin{equation}
\label{eqn:basic-likelihood-integral-y-only-on-theta-repeat}
  \begin{aligned}
    p(y|\theta) = \prod_{i=1}^n p(y_i|\theta).  
  \end{aligned}
\end{equation}

Then, the natural logarithm of this distribution is given as

\begin{equation}
\label{eqn:log-basic-likelihood-integral-y-only-on-theta}
  \begin{aligned}
    \ln {p(y|\theta)} = \ln {\prod_{i=1}^n p(y_i|\theta)} = \sum_{i=1}^n \ln {p(y_i|\theta)}.  
  \end{aligned}
\end{equation}

We \textbf{do not have agreement} between the formulation given in Eqn.~\ref{eqn:basic-free-energy-Beal-H-only} and that given in Eqn.~\ref{eqn:simple-H-3} together with Eqn.~\ref{eqn:H-in-terms-of-Q}.

Moving on, we now turn our attention to the derivation for the free energy as introduced in Beal \cite{Beal_2003_Variational-algorithm-approx-Bayes-inference}.

Beal introduces a formulation for the log likelihood in his Eqns. 2.12 - 2.16, reproduced here:

\begin{equation}
\label{eqn:basic-free-energy-Beal}
  \begin{aligned}
    L(\theta) 
&    = \sum_{i=1} ln \int dx_i \: p(x_i,y_i|\theta) \\
&    =  \sum_{i=1} ln \int dx_i \: q_{x_i}(x_i) \: \frac { p(x_i,y_i|\theta)} {q_{x_i}{(x_i})}  \\
&    \geq  \sum_{i=1} \int dx_i \: q_{x_i}(x_i) \: ln \: \frac { p(x_i,y_i|\theta)} {q_{x_i}{(x_i})} \\
&    =  \sum_{i=1} \left( \int dx_i \: q_{x_i}(x_i) \: ln  \: p(x_i,y_i|\theta) - \int dx_i \:  q_{x_i} (x_i) \: ln \: q_{x_i}(x_i) \right) \\
&    =  \sum_{i=1} \left( \int dx_i \: q_{x_i}(x_i) \: ln  \: p(x_i,y_i|\theta) \right)  - \int dx_i \:  q_{x_i} (x_i) \: ln \: q_{x_i}(x_i) \\
& \equiv F(q_{x_1}{(x_1}),..., q_{x_n}{(x_n}), \theta) 
.    
  \end{aligned}
\end{equation} 

Beal notes that $F(q_x(x),\theta)$ is a lower bound on $L(\theta)$ and is a functional of the free
distributions $q_{x_i}(x_i)$ and of $\theta $ (the dependence on $y$ is left implicit). The inequality introduced in the third expression makes use of \textit{Jensen's inequality}.

Beal notes: ``Defining the energy of a global configuration $(x, y)$ ...
the lower bound $F(q_x(x), \theta)\leq L(\theta)$ is the negative of a quantity known in statistical physics as the free energy: the expected energy under $q_x(x)$ minus the entropy of $q_x(x)$ (Feynman, 1972; Neal and Hinton, 1998)'' \cite{Feynman_1972_Statistical-Mechanics, Hinton-van-Camp_1993_Keeping-neural-networks-simple}.

Thus, both $F(q_x(x),\theta)$ and $L(\theta)$ refer to free energy formalization, however, $L(\theta)$ can be greater than $F(q_x(x),\theta)$. Alternatively, we can say that $F(q_x(x),\theta)$ is a \textit{lower bound} on $L(\theta)$.

Essentially (in Beal's conception; Friston's is reversed), we're saying that the free energy of the model is going to be lower than the free energy of the system being modeled; $L \geq F$. If we are going to improve the accuracy of the model, we will be bringing $F(q_x(x),\theta)$ towards $L(\theta)$. 

Beal further notes that \textbf{$F(q_x(x),\theta)$ is the negative} of what is known, in statistical thermodynamics, as the \textit{free energy} of a system, which is the \textit{expected energy} ($H$) under $q_{x}(x)$ minus the entropy of $q_{x}(x)$. Thus, when we shift to the notation of Friston (op.cit.), we will reverse the signs on all of the terms on the right-hand-side of Eqn.~\ref{eqn:basic-free-energy-Beal} (second-to-last line of the equation), leading up to Beal's definition of   $F(q_x(x),\theta)$.

According to this understanding, and changing the signs of the terms to deal with Beal's use of $F(q_x(x),\theta)$ as the \textbf{negative} of the free energy, we have the expected energy (or enthalpy) of a system $H$ is given as

\begin{equation}
\label{eqn:basic-eqn-enthalpy-H}
  \begin{aligned}
    H =  - \sum_{i=1}  \int dx_i \: q_{x_i}(x_i) \: ln  \: p(x_i,y_i|\theta),    
  \end{aligned}
\end{equation} 

\noindent
and

\begin{equation}
\label{eqn:basic-eqn-entropy-S}
  \begin{aligned}
    S =  - \sum_{i=1}  \int dx_i \:  q_{x_i} (x_i) \: ln \: q_{x_i}(x_i).    
  \end{aligned}
\end{equation}

The definition for the entropy of a system is given as 

\begin{equation}
\label{eqn:really-basic-eqn-entropy-S}
  \begin{aligned}
    S = - k_{\beta} \: ln \: \Omega.    
  \end{aligned}
\end{equation} 

An alternative formulation is

\begin{equation}
\label{eqn:really-basic-eqn-entropy-S-2}
  \begin{aligned}
    S = - k_{\beta} \: \langle ln \: P \rangle,    
  \end{aligned}
\end{equation} 

\noindent
where $P$ is the probability distribution over all states. Thus, $S$ is the expectation of the probability of the variable for which various states are possible; i.e., this correlates directly with Eqn.~\ref{eqn:basic-eqn-entropy-S}, and 
where $\Omega$ is the \textit{grand partition function}, and $k_{\beta}$ is Boltzmann's constant.

From Feynman (1972) \cite{Feynman_1972_Statistical-Mechanics}, Eqn. 1.1, that

\begin{equation}
\label{eqn:Feynman-Q}
  \begin{aligned}
     Q =  \sum_{n}  \:  \exp (-E_n/{kT} ) \equiv \exp (-F/{kT} ).    
  \end{aligned}
\end{equation}

From Feynman (1972) we have the definition for $F$, the Helmholtz free energy, as

\begin{equation}
\label{eqn:Feynman-free-energy}
  \begin{aligned}
    F =  -kT \ln Q = - kT \ln \left( \sum_{n}  \:  \exp (-E_n/{kT} ) \right)    
  \end{aligned}
\end{equation}
 
\noindent and also  (Eqn. 1.3)
 
\begin{equation}
\label{eqn:Feynman-entropy}
  \begin{aligned}
    S =  - k   \sum_{n}  \:  P_n \ln \: P_n,     
  \end{aligned}
\end{equation}

\noindent where (Eqn. 1.4)

\begin{equation}
\label{eqn:Feynman-prob-energy}
  \begin{aligned}
    P_n =  \frac{1}{Q} \:  \exp (-E_n/{kT} ).     
  \end{aligned}
\end{equation}

Further, the average energy per unit, $U$ (also known as the enthalpy), is given as (Eqn. 1.7)

\begin{equation}
\label{eqn:Feynman-enthalpy}
  \begin{aligned}
    U =  \frac{1}{Q} \:   \sum_{n}  \:  E_n \: \exp (-E_n/{kT} ),     
  \end{aligned}
\end{equation}

\noindent which can be rewritten as

\begin{equation}
\label{eqn:Feynman-enthalpy-rewritten}
  \begin{aligned}
    U =    \sum_{n}  \:  E_n \: P_n.     
  \end{aligned}  
\end{equation}

Essentially, Eqn.~\ref{eqn:Feynman-enthalpy-rewritten} states that the enthalpy of a system (the total energy of the system), is simply the sum of the energy-per-state, multiplied by the probability that a unit will be in that state, where the whole is summed over all the units in the system. 

This derivation by Feynman \cite{Feynman_1972_Statistical-Mechanics} is perhaps a bit more intuitive than that put forth in Appendix A, and may be preferred.

%
\section{Appendix: The Cluster Variation Method}
\label{sec:Appendix-cluster-variation-method}
%

  \renewcommand{\theequation}{C-\arabic{equation}}
  \setcounter{equation}{0}  

This Appendix provides a very brief overview of the cluster variation method (CVM), specifically the 2-D CVM. In particular, it provides: 

\begin{enumerate}
\item Notion of the \textit{configuration variables}, together with illustrations of how they appear within a 2-D CVM grid, 
\item The\textit{ free energy equation} for the 2-D CVM, along with the equations for the equilibrium values for the different configuration variables, which are analytically available when there is an equiprobable distribution of nodes into states \textbf{A} and \textbf{B}; i.e., $x_1 = x_2 = 0.5$, and
\item A very quick sketch of the actual \textit{free energy minimization protocol}, which defines how an actual at-equilibrium 2-D CVM grid configuration can be obtained, together with all associated configuration variable values and thermodynamic values for the system. 
\end{enumerate}

The goal of this Appendix is to provide just enough information to make Section~\ref{sec:var-FE-new-comput-engine} more readable for those who wish to understand the potential role of a 2-D CVM system in a computational system embodying a variational Bayes approach, within the context used here of separating the representational system from the external one via a Markov blanket. 

For those wishing to pursue the 2-D CVM in more depth, the most accessible starting place is provided in Maren (2016) \cite{Maren_2016_CVM-primer-neurosci}. This overviews the CVM approach for both 1-D and 2-D systems. It also suggests how, in particular, a 2-D CVM approach can potentially relate to modeling neural activation topographies, and can also fit in with descriptions of statistical mechanics in the brain. 

The CVM was first developed by Kikuchi \cite{Kikuchi_1951_Theory-coop-phenomena}, and then jointly by Kikuchi and Brush \cite{Kikuchi-Brush_1967_Improv-CVM}. Kikuchi and Brush presented an analytic solution for both the 1-D and 2-D systems for the equiprobable distribution case. They did not provide details for this analytic solution. The derivation steps for the 1-D case are in Maren (2014a) \cite{AJMaren-TR2014-002}, and for the 2-D case are in Maren (2014b, 2019) \cite{AJMaren-TR2014-003, AJMaren-TR2019-02-2D-CVM-FE-Fund-and-Prag}.

%
\subsection{Configuration Variables }
\label{subsec:configuration-variables}
%

The cluster variation method, introduced by Kikuchi \cite{Kikuchi_1951_Theory-coop-phenomena} and refined by Kikuchi and Brush \cite{Kikuchi-Brush_1967_Improv-CVM}, uses an entropy term that includes not only the distribution across simple ``on'' and ``off'' states, but also the distribution into local patterns, or \textit{configurations}, as illustrated in the following Figure~\ref{fig:Config-var-weights_v3_crppd_2017-05-17}.

A 2-D CVM is characterized by a set of \textit{configuration variables}, which collectively represent single unit, pairwise combination, and triplet values. The configuration variables are denoted as: 

\begin{itemize}
\setlength{\itemsep}{1pt}
\item $x_i$ - Single units, 
\item $y_i$ - Nearest-neighbor pairs, 
\item $w_i$ - Next-nearest-neighbor pairs, and 
\item $z_i$ - Triplets. 
\end{itemize}

The degeneracy factors $\beta_i$ and $\gamma_i$ (number of ways of constructing a given configuration variable) are shown in the following Figure~\ref{fig:Config-var-weights_v3_crppd_2017-05-17}; $\beta_2 = 2$ for both $y_2$ and $w_2$,  as $y_2$ and $w_2$ can be constructed as either \textbf{A}-\textbf{B} or as \textbf{B}-\textbf{A} for $y_2$, or as \textbf{B}-~-\textbf{A} or as \textbf{A}-~-\textbf{B} for $w_2$. All other degeneracy factors for the  $y_i$ and $w_i$ configuration variables are set to 1.

\begin{figure}[ht]
  \centering
  \fbox{
  \rule[-.5cm]{0cm}{4cm}\rule[-.5cm]{0cm}{0cm}	
  \includegraphics [trim=0.0cm 0cm 0.0cm 0cm, clip=true,   width=0.95\linewidth]{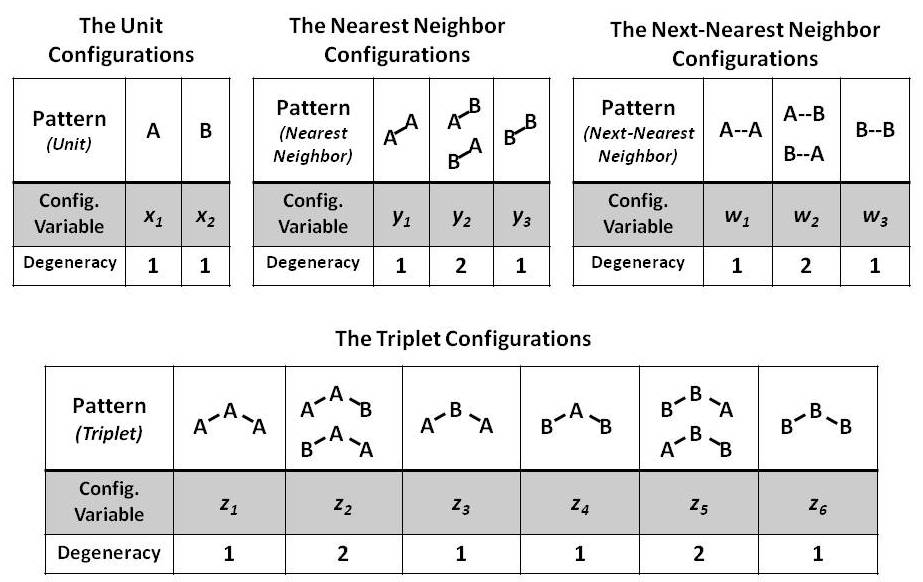}}
  \vspace{3mm} 
  \caption{Illustration of the \textit{configuration variables} for the cluster variation method, showing the ways in which the configuration variables $y_i$, $w_i$, and $z_i$ can be constructed, together with their degeneracy factors $\beta_i$ and $\gamma_i$.}   
\label{fig:Config-var-weights_v3_crppd_2017-05-17}
\end{figure}
\vspace{3mm} 

Notice also that within Figure~\ref{fig:Config-var-weights_v3_crppd_2017-05-17}, the triplets $z_2$ and $z_5$ have two possible configurations each: \textbf{A}-\textbf{A}-\textbf{B} and \textbf{B}-\textbf{A}-\textbf{A} for $z_2$, and \textbf{B}-\textbf{B}-\textbf{A} and \textbf{A}-\textbf{B}-\textbf{B} for $z_5$. This means that there is a degeneracy factor of 2 for each of the $z_2$ and $z_5$ triplets.  This gives us $\gamma_2 = \gamma_5 = 2$ (for the triplets), as there are two ways each for constructing the triplets $z_2$ and $z_5$. The remaining degeneracy factors for the triplet configuration variables are set to 1. 

%
\subsection{Free Energy for the 2-D CVM}
\label{subsec:free-energy-2D-CVM}
%

The analytic solution for the case where $x_1 = x_2 = 0.5$ can be found when we are using the full interaction enthalpy term of $\varepsilon_1*(2y_2-y1-y3)$. This solution is presented in Maren (2019) \cite{AJMaren-TR2019-02-2D-CVM-FE-Fund-and-Prag}.

The free energy equation for a 2-D CVM system, including configuration variables in the entropy term, is

\begin{equation}
\label{Bar-G-2-D-basic-eqn}
  \begin{aligned}
\bar{G}_{2-D} = G_{2-D}/N = \\
  & \varepsilon_1(-z_1+z_3+z_4-z_6) \\
- & \Bigg[ 2 \sum\limits_{i=1}^3 \beta_i Lf(y_i)) 
          + \sum\limits_{i=1}^3 \beta_i Lf(w_i)) 
          - \sum\limits_{i=1}^2 \beta_i Lf(x_i)) 
          - 2 \sum\limits_{i=1}^6 \gamma_i Lf(z_i) \Bigg]\\
+ & \mu (1-\sum\limits_{i=1}^6 \gamma_i  z_i )+4 
\lambda (z_3+z_5-z_2-z_4)
  \end{aligned}
\end{equation}

\noindent
where $\mu$ and $\lambda$ are Lagrange multipliers, and  we have set $k_{\beta}T = 1$. 

The single enthalpy parameter here is $\varepsilon_1$, with the enthalpy parameter for unit activation implicitly set to zero, as the intention has been to solve the above equation for an analytic solution, which is possible only in the case where $x_1 = x_2 = 0.5$, meaning that the per-unit enthalpy activation parameter $\varepsilon_0 = 0$.

When we use $\varepsilon_1(2y_2-y_1-y_3) = \varepsilon_1(-z_1+z_3+z_4-z_6)$ as the enthalpy expression (as is done in the previous equation), we can obtain an analytic solution for each of the configuration variables. For example, we find the expression for $z_3$ in terms of $h$ as 

\begin{equation}
\label{Eqn:z3-analyt1-current-approach-solution}
  z_3=\frac{(h-3)(h+1)}{8[h^2-6h+1]}. 
\end{equation}\\
\vspace{-12pt}

(\textit{Note:} See Maren (2019) \cite{AJMaren-TR2019-02-2D-CVM-FE-Fund-and-Prag} for full derivations. This corresponds with Eqn. (I.25) in Kikuchi and Brush (1967) \cite{Kikuchi-Brush_1967_Improv-CVM}.)

Similar expressions can be obtained for the remaining configuration variables. 

The experimentally-generated results from probabilistically-generated data sets correspond to the analytic results in the neighborhood of $\varepsilon_0=0$. The reason that the range is so limited is that the analytic solution makes use of equivalence relations as expressed above. 

The resulting solution of Eqn.~\ref{Eqn:z3-analyt1-current-approach-solution} clearly has a divergence in it, due to the quadratic expression in the denominator. There are two solutions to the quadratic expression. We are interested in the case where the value of $h>1$ indicates that $\varepsilon_1 >0$, which is the case where the interaction enthalpy favors like-near-like interactions, or some degree of gathering of similar units into clusters. This means that we expect that the computational results would differ from the analytic as $h$ approaches the divergence point. 

The divergence in the analytic solution is an artifact. However, it does indicate that for larger \textit{h-values}, the analytic solution will not be accurate. Thus, for high \textit{h-values}, we take the analytic solution simply as a starting point, and invoke a \textbf{\textit{protocol}} for determining the correct configuration variable values associated with a given \textit{h-value}, as described in Maren \cite{AJMaren-TR2019-02-2D-CVM-FE-Fund-and-Prag}.

%
\subsection{Free Energy Minimization Protocol for the 2-D CVM}
\label{subsec:protocol}
%

The following describes an early protocol for finding an \textit{h-value} that is in proximity to the \textit{h-value} that would be a ``best fit for a given 2-D CVM grid. It is included here largely for historical reasons (Maren 2021, 2019) \cite{AJMaren-2021-2D-CVM-Topography, AJMaren-TR2019-02-2D-CVM-FE-Fund-and-Prag}.

More recently, Maren (2022) has devised a new protocol that uses a new divergence measure, the \textit{Kikuchi-Maren divergence}, for identifying the activation enthalpy parameter $\varepsilon_1$ that provides a ``best fit'' for characterizing a given 2-D CVM topography. This can then be used to bring that topography to a free energy minimum \cite{AJMaren-2022-Variational-Approach-2D-CVM}.

An initial 2-D CVM grid, whether manually designed or randomly-generated, is typically not at equilibrium. More specifically, the various configuration variable values will typically correspond to different \textit{h-values}, where the \textit{h-value} or simply, \textit{h} is defined as $h=exp(2\varepsilon_1)$. 

Each \textit{h-value} indicates a unique free energy minimum solution. Thus, we need to take the initial grid through a free energy minimization process, in order to obtain a grid where the various configuration variables all correspond to a single free energy minimum. This requires using a \textit{free energy minimization protocol}. 

This protocol takes a computational approach, so that given an initial 2-D CVM grid pattern, the steps are to:

\begin{enumerate}
\item Obtain an initial estimate for a candidate \textit{h-value},
\item Determine the associated thermodynamic values for that particular \textit{h-value} estimate and current set of configuration values, and
\item Adjust the node activations in order to minimize the free energy for that given \textit{h-value}, yielding a new grid configuration and associated set of configuration variables and thermodynamic values. 
\end{enumerate}

See Maren \cite{AJMaren-TR2018-001v2-V-and-V} for the code verification and validation documentation. Code releases are being provided courtesy of Themesis, Inc. and are available at the Themesis GitHub repository \cite{Themesis-GitHub}. 

As a preliminary step, a computer program is used to obtain the actual counts for each of the different configuration variables. The next step is to find the corresponding \textit{h-values} for each of three different configuration variables; $z_1$, $z_3$, and $y_2$. These are selected because, taken together, they are reasonably descriptive of the grid topography: 

\begin{itemize}
\item $z_1$ - \textbf{A}-\textbf{A}-\textbf{A} triplets; indicates the relative fraction of \textbf{A} units that are included in ``islands'' or ``land masses''; this also (indirectly) indicates the compactness of these masses,
\item $z_3$ - \textbf{A}-\textbf{B}-\textbf{A} triplets; indicates the relative fraction of \textbf{A} units that are involved in a ``jagged'' border (one that involves irregular protrusions of \textbf{A} into a \textbf{B} space), or the presence of one or more ``rivers'' of \textbf{B} units extending into landmass(es) of \textbf{A} units, and
\item $y_2$ - \textbf{A}-\textbf{B} nearest-neighbor pairs; indicates the relative extent to which the \textbf{A} units are distributed among the surrounding \textbf{B} units. 
\end{itemize}

We briefly describe the first protocol used. Initially, each of these configuration variables will correspond to a different \textit{h-value}. (This corresponding \textit{h-value} can be found graphically, via a look-up table, or by extrapolation, using Eqn.~\ref{Eqn:z3-analyt1-current-approach-solution} and similar equations.) This gives us a set of (typically) three \textit{h-values}. 

According to this initial protocol, we can determine which \textit{h-value} to use via any number of means; it is reasonable to take an average or mean of the \textit{h-values} corresponding to the configuration variable values. 

This \textit{h-value} then is used to define the corresponding $\varepsilon_1$ in a program that will iteratively modify the grid configuration. (This is for the case where $x_1 = x_2 = 0.5$, so that $\varepsilon_0 = 0$.) 

More specifically, the program will randomly select two units, and ascertain that the first is in the \textbf{A} state and the other is in the \textbf{B} state. (Random selection continues until one of each has been found.) The program then ``swaps'' the two nodes' states; that which was \textbf{A} becomes \textbf{B} and vice versa. The program then computes the new set of configuration variables and the corresponding new thermodynamic values. If the free energy is reduced, the swap is kept; if not, it is reversed. The program continues this for a predetermined number of trials. 

Graphical display of the free energy values through this process shows that the free energy typically drops relatively fast, and then is impervious to future node swaps. 

This is, of course, an extremely heavy-handed and simplistic approach. It does not take into account any sense of what might be accomplished by intelligently selecting candidate nodes for which a ``flip'' in activation would have the greatest free energy impact. Such a more sophisticated strategy would require an object-oriented approach, and the current program is simplistic in terms of node representations. 

However, these initial and exploratory experiments have produced interesting topologies. For example, as shown in the body of this article, a free energy-minimized grid topography will often have ``spider legs'' of \textbf{A} units connecting various islands and landmasses of \textbf{A}. Studies of free energy-minimized topographies are in progress. 

In the more recent protocol (Maren, 2022) \cite{AJMaren-2022-Variational-Approach-2D-CVM}, we similarly obtain an intiial potential range of \textit{h-values}. We then use the reverse Kikuchi-Maren divergence to identify the candidate \textit{h-value} that provides the smallest divergence between a (figurative) system with at-equilibrium configuration variable values as compared with the actual representational system. The \textit{h-value} that provides the smallest divergence is selected, and is used to provide a ``target'' system against which the reperesntational system is adjusted until it has configuration variable values that are as close as desired (within the limits of simple node-activation swapping) with the target values. This is a situation in which Action Perception Divergence (APD) (Hafner et al., 2020, rev. 2022) \cite{Hafner-et-al_2022_Action-perception-divergence} usefully provides a means for bringing the representational system through a ``phase space'' defined by the two control parameters $(\varepsilon_0, \varepsilon_1)$ and their associated configuration variable values.

%
\subsection{Application to a Variational Bayes Approach}
\label{subsec:Application}
%

The notion behind offering the 2-D CVM for the extension of variational Bayes into modeling a representational system rests on the premise that both the external and representational systems, composed of  $\tilde{\psi}$ and $\tilde{r}$ units respectively, can be expressed using a 2-D CVM. The elements for the representational system would be created by sampling areas of the external system; that is, the sensory elements $\tilde{s}$ of the Markov blanket would generate elements of $\tilde{r}$. 

It is reasonable that neither the external nor the representational systems would be completely at free energy equilibrium. In the case of the $\tilde{\Psi}$ system, this could reasonably be due to local influences and events, and also as the system can be differentially changing in response to various inputs. In the case of the representational system with $\tilde{r}$ units, this can be attributed to creating its unit activations via sampling from the $\tilde{\psi}$ units. 

The model $q$ is formed by bringing the representational system $\tilde{r}$ into free energy equilibrium. In this approach, we would actually go through a free energy minimization process, as that is literally feasible when we have a 2-D CVM grid. The resulting parameters $\theta$ are the set of $(\varepsilon_0, \varepsilon_1)$ values that describe the free energy-minimized resultant 2-D CVM grid. These two parameters then indicate the corresponding configuration variables, which define the nature of the grid's topography. 

While this, in itself, is somewhat interesting, the real value would lie in taking this kind of 2-D CVM grid into a computational engine. Inputs from an external source ($\tilde{s}$) can be used to generate certain activations across the grid. The free energy minimization process then modifies these activations. The resulting node activations can then be ``learned'' (using any number of neural network learning methods). 

What makes this process potentially interesting and useful is that we can then take one more step. We can create temporal persistence of a unit's activation, as a function of the degree to which it is central to an island or landmass of similar units. In short, we have a means for inducing temporal persistence that is dependent on a form of lateral interactions (i.e., neighborliness to units that are similarly in an active state). This means that a given unit's activation is now a function of two factors; the typical input from an external stimulus, and a lateral interaction. 

This method will be addressed more fully in subsequent works.


%
%


\end{document}